%% file: main.tex
\documentclass[twocolumn,letterpaper]{IEEEAerospaceCLS}  % only supports two-column, letterpaper format
\usepackage{amsmath}
\usepackage{amssymb}
\usepackage[]{graphicx}    % We use this package in this document
\newcommand{\ignore}[1]{}  % {} empty inside = %% comment
\usepackage{enumitem}
\usepackage{booktabs}
\usepackage{array}
\usepackage{siunitx}
\usepackage{tabularray}
\UseTblrLibrary{booktabs}
\usepackage[table]{xcolor}
\usepackage{placeins}

\graphicspath{{images/}}

\include{macros}

\begin{document}

\title{A physics-based sensor simulation environment for lunar ground operations}

\author{%
Nevindu M. Batagoda$^\dagger$\thanks{$\dagger$Equal contribution.}\\
Dept. of Mechanical Engineering\\
University of Wisconsin - Madison\\
batagoda@wisc.edu
\and 
Bo-Hsun Chen$^\dagger$\footnotemark[1]\\
Dept. of Computer Sciences\\
University of Wisconsin - Madison\\
bchen293@wisc.edu
\and
Harry Zhang\\
Dept. of Mechanical Engineering\\
University of Wisconsin - Madison\\
hzhang699@wisc.edu
\and
Radu Serban\\
Dept. of Mechanical Engineering\\
University of Wisconsin - Madison\\
serban@wisc.edu
\and
Dan Negrut\\
Dept. of Mechanical Engineering\\
University of Wisconsin - Madison\\
negrut@wisc.edu
%%%% IMPORTANT: Use the correct copyright information--IEEE, Crown, or U.S. government. %%%%%
\thanks{\footnotesize 979-8-3503-5597-0/25/$\$31.00$ \copyright2025 IEEE}      
}

\maketitle

\thispagestyle{plain}
\pagestyle{plain}

\maketitle

\thispagestyle{plain}
\pagestyle{plain}

\begin{abstract}
  This contribution reports on a software framework that uses physically-based rendering to simulate camera operation in lunar conditions. The focus is on generating synthetic images qualitatively similar to those produced by an actual camera operating on a vehicle traversing and/or actively interacting with lunar terrain, e.g., for construction operations. The highlights of this simulator are its ability to capture ($i$) light transport in lunar conditions and ($ii$) artifacts related to the vehicle-terrain interaction, which might include dust formation and transport. The simulation infrastructure is built within an in-house developed physics engine called Chrono, which simulates the dynamics of the deformable terrain--vehicle interaction, as well as fallout of this interaction. The Chrono::Sensor camera model draws on ray tracing and Hapke Photometric Functions. We analyze the performance of the simulator using two virtual experiments featuring digital twins of NASA's VIPER rover navigating a lunar environment, and of the NASA's RASSOR excavator engaged into a digging operation. The sensor simulation solution presented can be used for the design and testing of perception algorithms, or as a component of in-silico experiments that pertain to large lunar operations, e.g., traversability, construction tasks.  
\end{abstract} 

\tableofcontents

%%%%%%%%%%%%%%%%%%%%%%%%%%%%%%%%%%%%%%
\section{Introduction}
\label{sec:intro}
\input{sections/introduction}
%%%%%%%%%%%%%%%%%%%%%%%%%%%%%%%%%%%%%%

%%%%%%%%%%%%%%%%%%%%%%%%%%%%%%%%%%%%%%
\section{Lunar Simulation Ecosystem}
\label{sec:ecosystem}
\input{sections/ecosystem}
%%%%%%%%%%%%%%%%%%%%%%%%%%%%%%%%%%%%%%

%%%%%%%%%%%%%%%%%%%%%%%%%%%%%%%%%%%%%%
\section{Camera Simulator for Lunar Conditions}
\label{sec:cameraSim}
\input{sections/camerasim}

%%%%%%%%%%%%%%%%%%%%%%%%%%%%%%%%%%%%%%

%%%%%%%%%%%%%%%%%%%%%%%%%%%%%%%%%%%%%%
\section{Demonstrations of the Technology}
\label{sec:experiments}
\input{sections/experiments}
\input{sections/experiments_bohsun}
\input{sections/experiments_YOLO}
%%%%%%%%%%%%%%%%%%%%%%%%%%%%%%%%%%%%%%

%%%%%%%%%%%%%%%%%%%%%%%%%%%%%%%%%%%%%%
\section{Limitations}
\label{sec:limitations}
\input{sections/limitations}
%%%%%%%%%%%%%%%%%%%%%%%%%%%%%%%%%%%%%%

%%%%%%%%%%%%%%%%%%%%%%%%%%%%%%%%%%%%%%
\section{Conclusions}
\label{sec:conclusions}
\input{sections/conclusions}

%%%%%%%%%%%%%%%%%%%%%%%%%%%%%%%%%%%%%%

%%%%%%%%%%%%%%%%%%%%%%%%%%%%%%%%%%%%%%%%%%%%%%%%%%%%%%%%%%%%%%%%%%%%%%%%%%%%%%%%%%%%%%%%%%%%%%%%%%%%%%
\acknowledgments
This work has been carried out with support from NASA project 80NSSC24CA030 and NSF projects CMMI2153855 and OAC2209791. The authors are very grateful for this funding. We would like to thank the following individuals from the University of Wisconsin-Madison: Luning Bakke for providing the RASSOR image, and Keshav Pachipala and Tony Adriansen for their valuable feedback on an earlier draft of this manuscript. We are also indebted to the foundational work of Dr. Asher Elmquist during his time as a student at the University of Wisconsin-Madison, which laid the groundwork for Chrono::Sensor. Finally, we are grateful to Arno Rogg of NASA Ames, who provided insightful feedback during our discussions and has been very supportive of our efforts.

\FloatBarrier

%%%%%%%%%%%%%%%%%%%%%%%%%%%%%%%%%%%%%%%%%%%%%%%%%%%%%%%%%%%%%%%%%%%%%%%%%%%%%%%%%%%%%%%%%%%%%%%%%%%%%%
\bibliographystyle{IEEEtran}
\bibliography{BibFiles/refsSensors,BibFiles/refsChronoSpecific,BibFiles/refsRobotics,BibFiles/refsSBELspecific,BibFiles/refsML-AI,BibFiles/refsMBS,BibFiles/refsCompSci,BibFiles/refsTerramech,BibFiles/refsFSI,BibFiles/refsDEM,BibFiles/refsGraphics,BibFiles/refsAutonomousVehicles}

\end{document}

%% file: macros.tex
\definecolor{arsenic}{rgb}{0.23, 0.27, 0.29}
\definecolor{charcoal}{rgb}{0.21, 0.27, 0.31}
\definecolor{hanblue}{rgb}{0.27, 0.42, 0.81}
\definecolor{blueofsorts}{rgb}{0.0, 0.53, 0.74}
\definecolor{awesome}{rgb}{1.0, 0.13,0.32}
\definecolor{darkgreen}{rgb}{0, .4,0}

\newcommand{\sbelNoIndentTitle}[1]{\noindent \textbf{{#1}}}

%% file: sections/introduction.tex
This work is focused on modeling sensors in the context of terramechanics applications, when one seeks to synthesize images used by the autonomy stack of the ground vehicle operating in deformable terrain conditions. We are interested in a holistic approach that captures the interplay between sensing, vehicle dynamics, and terramechanics. This topic is different than the important issue of simulating sensing for satellites in fly by operations or similar ``before touch-down'' remote sensing scenarios, which is discussed elsewhere, e.g., \cite{pangu2004,SISSPO2022,surrenderAirbus2018}. Specifically, the interest is in sensing at the scale of the vehicle and surrounding areas, i.e., two to three orders of magnitude less than the sizes associated with the scenarios discussed \cite{pangu2004,SISSPO2022,surrenderAirbus2018}.

Although the simulation infrastructure discussed is equally well applicable to terrestrial terramechanics, herein, the discussion is anchored by celestial body exploration, when producing synthetic images requires specialized techniques to address rendering difficulties posed by low light, long shadows, high dynamic range, the opposition effect, and minimal atmospheric light scattering. We describe a sensing framework that uses, as much as possible, physics-based simulation to capture in a principled way the process of image synthesis.  

\begin{figure}[htbp]
	\centering
	\includegraphics[width=\linewidth]{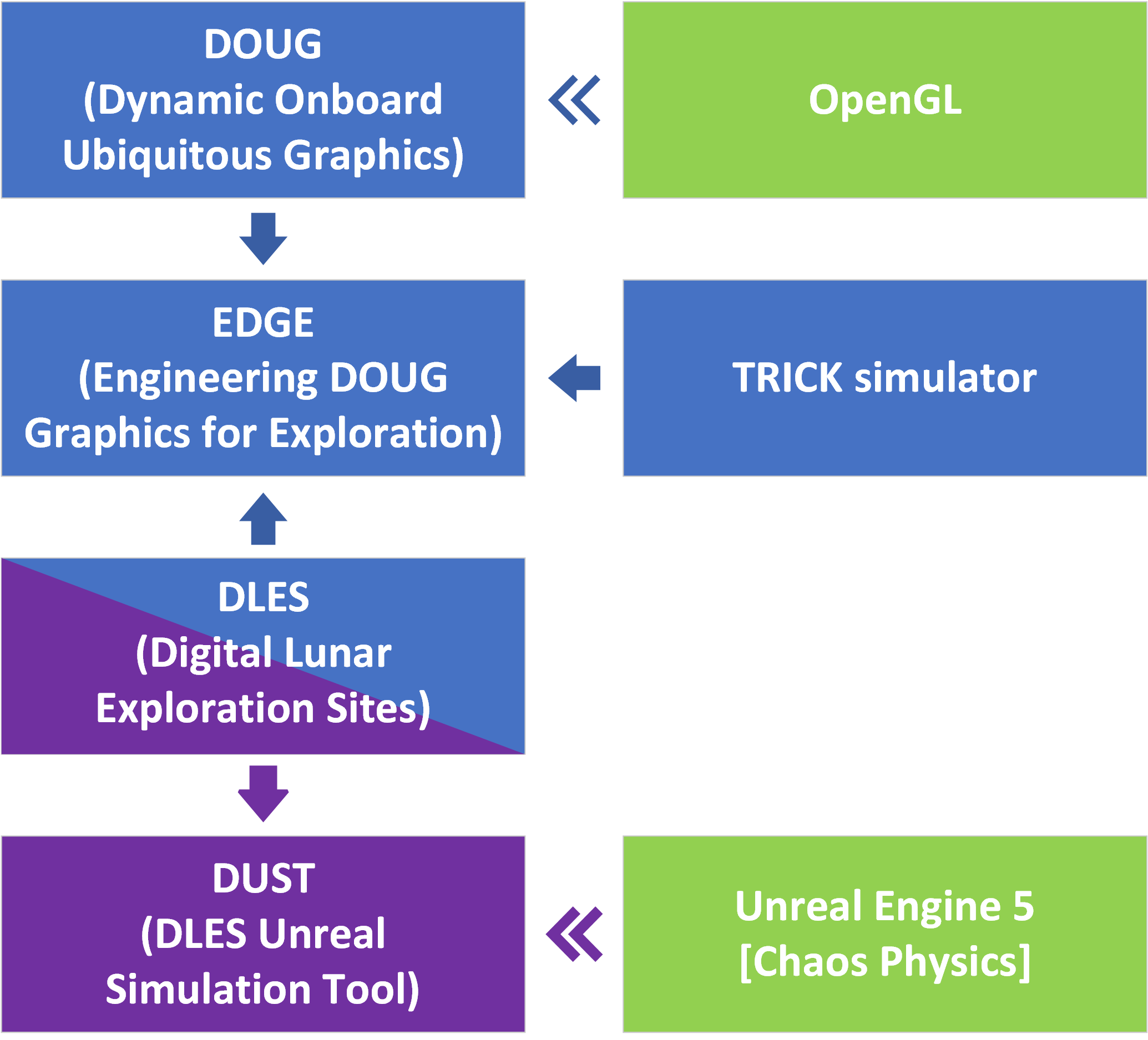}
	\caption{Interplay between several assets used in terrain simulation by NASA.}
	\label{fig:nasaSimAssets}
\end{figure}

Our simulation framework is similar in several respects to NASA simulators -- EDGE (Engineering DOUG Graphics for Exploration) and the newer DUST (DLES Unreal Simulation Tool). EDGE is NASA's proprietary real-time simulation and visualization platform that integrates graphics (via DOUG) and physics (via TRICK) to simulate and render space mission scenarios, particularly for lunar and planetary exploration (the acronyms are introduced in Fig.~\ref{fig:nasaSimAssets}). DOUG is a NASA produced real-time 3D graphics engine used for visualizing space missions, spacecraft operations, and astronaut training scenarios, providing the rendering framework for simulations like EDGE. Finally, unlike EDGE and DOUG which are proprietary, NASA's TRICK \cite{trick2016} is an open-source simulation environment that provides a framework for building and running physics-based simulations, often used to model spacecraft dynamics, robotic systems, and other mission-critical operations in real-time or faster than real-time. The final pillar of the EDGE platform is DLES (Digital Lunar Exploration Sites) \cite{dlesNASA2022}, a dataset developed by NASA that provides high-resolution, detailed topographic and environmental data of the lunar surface, particularly focused on the Lunar South Pole. It includes digital elevation models, terrain features like craters and rocks, and lighting conditions, and is used to support lunar mission planning, such as for NASA's Artemis program. The data from DLES is also used in a more recently developed DUST \cite{dust-NASA2023}, which is similar in its goals to EDGE but uses Unreal Engine 5 for rendering and simulation. For the latter, it does not rely on TRICK, but instead draws on the Chaos Physics dynamics engine. A simulator similar to DUST and EDGE is DARTS (Dynamics And Real-Time Simulation) \cite{DARTS}, which pairs with Iris \cite{irisDARTS2022} for sensor simulation. DARTS defines terrains through procedural methods, using a combination of Perlin noise, Voronoi noise, and other noise functions to generate multifractal patterns and complex terrains. Finally, a Gazebo-based simulator has been put together through a joint effort between NASA-Ames and Open Robotics that led to a Lunar rover simulator \cite{allan2019}. The physics engine used was Open Dynamics Engine (ODE) \cite{odeSmith2000}, which is a gaming engine in the vein of PhysX \cite{physxNVIDIA} and Chaos Physics. The simulation framework did not accommodate deformable terrain and instead used an empirical drawbar-pull coefficient vs. slip curve to account for slip phenomena. For graphics, it used Ogre3D \cite{ogre2002}, to handle the terrain's real-time rendering and shadowing effects. Modifications were made to Ogre3D's shadow mapping algorithm to improve shadow quality and optimize rendering for lunar terrains. Finally, terrains are generated using a combination of  Digital Elevation Models (DEMs) and procedural techniques to create high-resolution, realistic environments. To increase the resolution of the DEMs, the simulation employs fractal synthesis techniques. The placement of craters and rocks follows size-frequency distribution models derived from lunar observations. Finally, a custom GLSL shader was developed to model the reflective properties of lunar regolith, enhancing realism by simulating the unique lighting and reflectance conditions of the lunar surface such as long shadows and the opposition effect.

The EDGE, DUST, Gazebo-based, and DARTS platforms, developed at NASA JSC, NASA JSC, NASA Ames, and JPL, respectively, are not publicly available. The same is true for the URSim platform \cite{sewtz2022ursim} developed by Germany's Deutsches Zentrum f\"ur Luft- und Raumfahrt (DLR). URSim is conceptually similar to DUST, as it also relies on Unreal Engine, though it uses an older version—4.0 \cite{software-unreal}, and therefore utilizes NVIDIA's PhysX dynamics engine. URSim offers photo-realistic visual and physics support in real-time, and it is used to test and evaluate full robotic systems by allowing the investigation of the perception-action interplay. 

From the commercial world, NVIDIA's IsaacSIM \cite{isaacNVIDIA} is an integrated platform (in the sense that it offers both dynamics and sensor simulation) making inroads into robotics simulation. While freely available, it is not open source. Although both PhysX and IsaacSim are NVIDIA simulators, the former is targeted to game development, while the latter is positioned as an engineering-grade simulator. IsaacSim does not natively support lunar terrain generation, an aspect that has been recently investigated in \cite{kamoharaGrouserDefTerrain2024}. As for an effort that seeks to produce a simulation platforms based on IsaacSim, the reader is referred to \cite{richard-photoRealisticLunar2024}.

While EDGE, DUST, DARTS, URSim, and IsaacSIM combine the image synthesis needed in perception with a dynamics engine for physics simulation, GUISS (Graphical User Interface Simulation Software) \cite{icyMoonsBhaskara2024} is exclusively concerned with simulating sensor operation in icy environments. Camera sensing is emulated using Blender Cycles \cite{blender}. In this context, since it does not embed a dynamics engine, GUISS is similar to DLR's Oaysis \cite{muller2021photorealistic}, which targets real-time terrain rendering, and BlenderProc \cite{denninger2023blenderproc2}, which provides a procedural pipeline aimed at generating photorealistic renderings and semantic datasets, primarily used for training neural networks in tasks like computer vision and robotic perception. A similar line of work, though unrelated to extraterrestrial exploration, is reported in \cite{lyu2022validation}, where the focus is on high-fidelity simulation of image registration by a camera sensor or the human eye. The ISET toolbox enabled the authors to quantify the effects of camera parameter variations (including pixel sizes and color filters) and image processing operations on perception tasks \cite{liu2020neural}. This level of sensor simulation accuracy in ISET is also targeted to prototype image acquisition systems for autonomous driving \cite{blasinski2018optimizing}.

This contribution highlights how Chrono \cite{chronoOverview2016,chronoAPIWebSite} enables the simulation of autonomous and human-operated ground vehicles in extraterrestrial environments. At a high level, Chrono addresses the need for an open-source, publicly available simulator \cite{projectChronoGithub} capable of handling sensing in deformable terrain scenarios. Specifically, the simulator: ($i$) can replicate both active and passive light sensing processes, aiding in the training and testing of perception algorithms; ($ii$) can be utilized in the mechanical design of robots; and ($iii$) for certain applications that allow for real-time simulation, it can integrate with ROS-based autonomy stacks. However, regarding ($iii$), due to the complexity of terramechanics simulations, certain scenarios, such as digging, bulldozing, or tracked vehicles on deformable terrain, do not run in real-time in Chrono. Additionally, Chrono supports simultaneous hardware-in-the-loop (HIL) and software-in-the-loop (SIL) testing, allowing the simulator to test the physical chip (hardware) running the actual robot autonomy stack (software), while simulating the sensing processes and vehicle-environment dynamics.

Chrono offers functionality similar to that provided by EDGE, DUST, DARTS, and URSim. Chrono's key strength lies in its ability to simulate terramechanics in a principled manner, while simultaneously sensing the vehicle-environment interaction: the agent alters the environment, and the environment shapes the vehicle's response through the agency of a human operator or an autonomy stack. To the best of our knowledge, there is no other physics-based, \textit{open-source} terramechanics simulator that has the ability to sense in real time or close to real time the deformation of the soil as various implements, e.g., blades, grousers, buckets, drills, interact with it. Chrono has been or is currently being used in several NASA-sponsored projects, such as the VIPER mission, the RASSOR excavator \cite{RASSOR2013}, Moon Racer, and the Moon Ranger project at Carnegie Mellon University. 

\begin{figure}[h]
	\centering
	\includegraphics[width=0.45\textwidth]{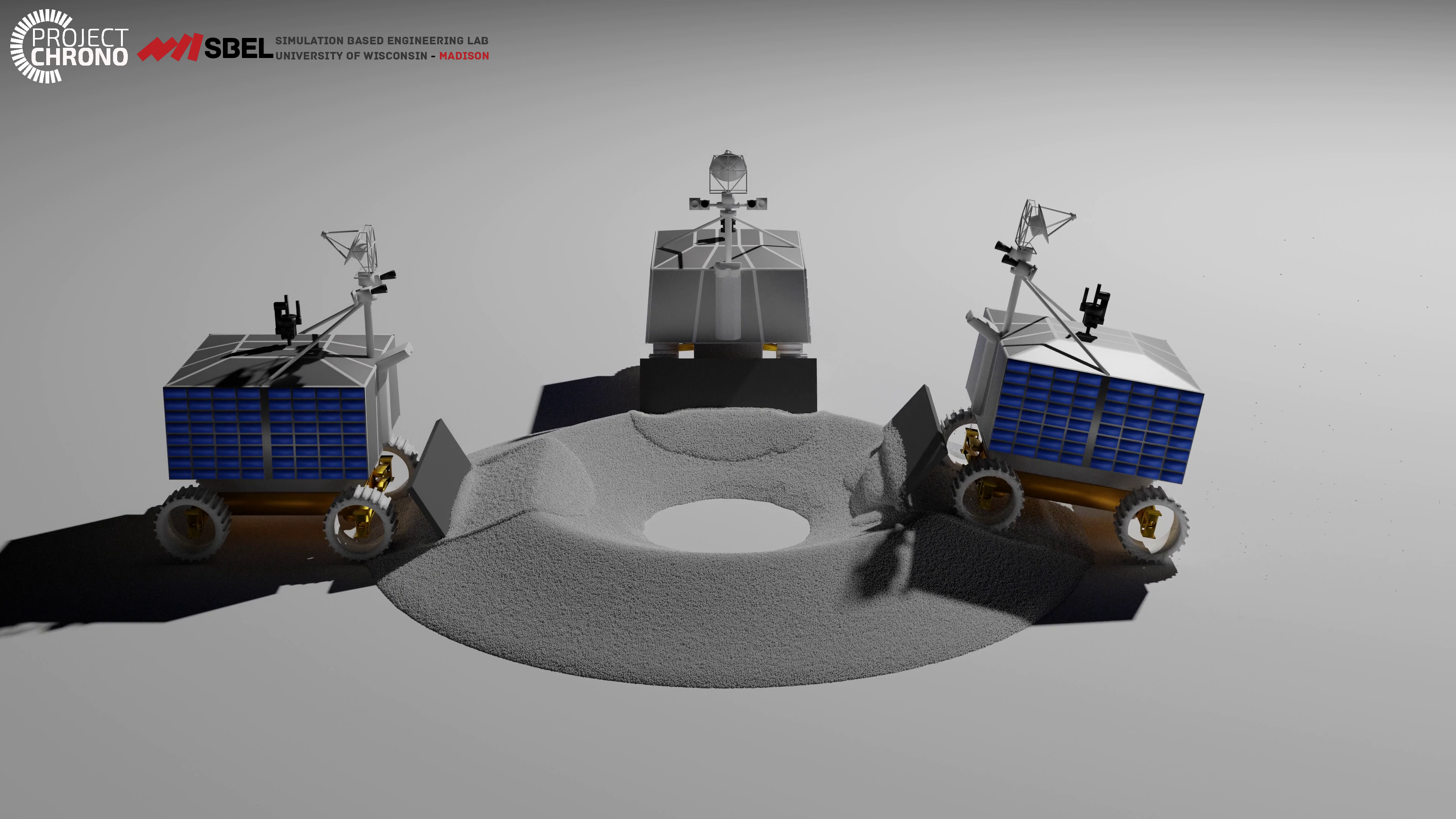}  % First image
	\includegraphics[width=0.45\textwidth]{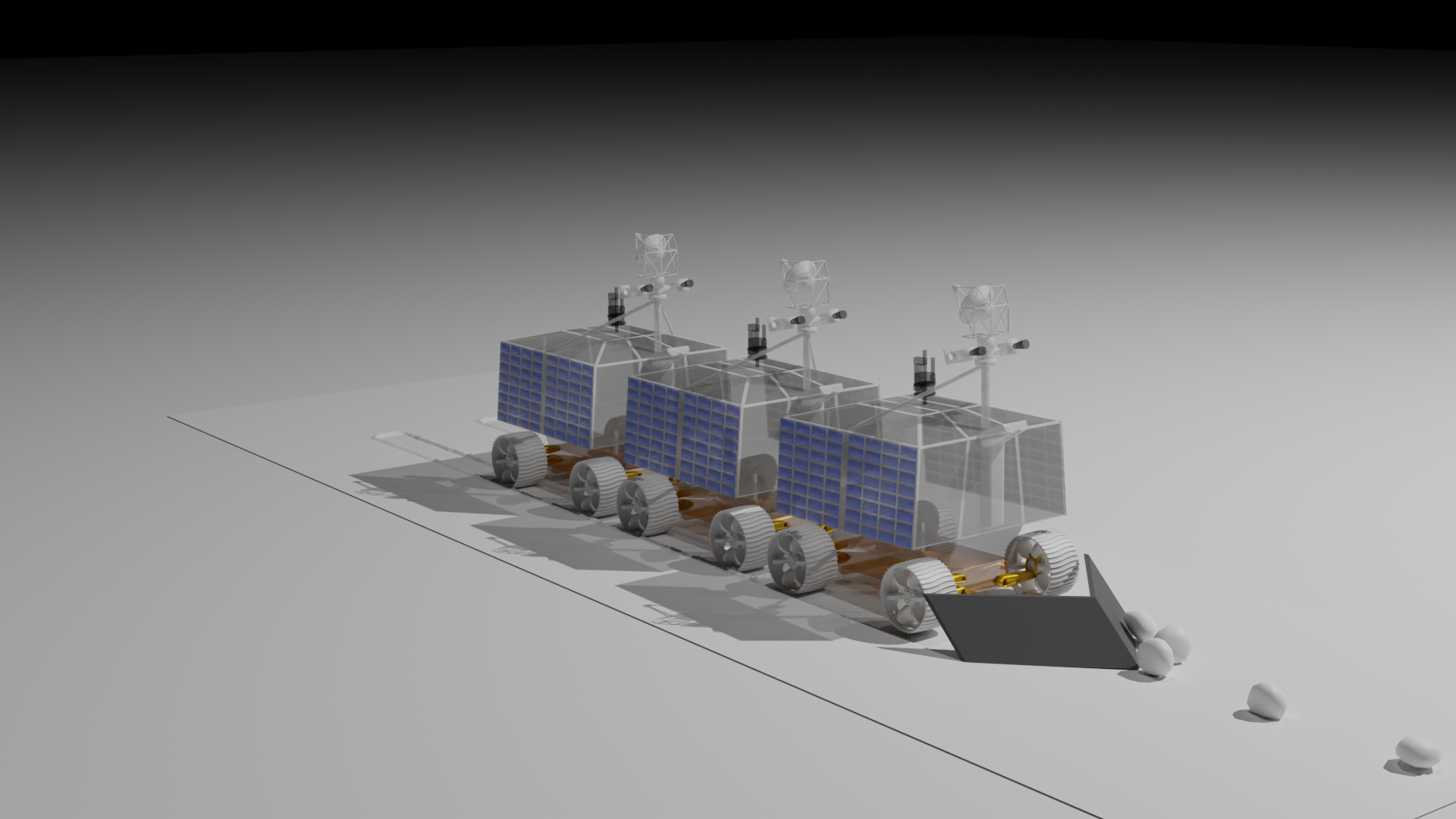}  % Second image
	\caption{Two Chrono simulation scenarios involving VIPER replicas involved in construction-type operations.}
	\label{fig:vipersAtWorks}
\end{figure}

This contribution provides an overview of Chrono's functionality relevant to ground vehicle-enabled extraterrestrial exploration, with an emphasis on camera sensing. In Section \ref{sec:ecosystem}, we discuss vehicle, terramechanics, and dust modeling aspects, while also briefly touching on other simulation capabilities. Section \ref{sec:cameraSim} explores the camera simulation and rendering process in greater detail, highlighting the use of physically-based rendering (PBR) and ray tracing as powered by NVIDIA's OptiX library \cite{optixNVIDIA}. Section \ref{sec:experiments} presents several virtual experiments conducted with Chrono. We include a section that discusses limitations of the simulator, and close with a conclusions section.

%% file: sections/ecosystem.tex
\sbelNoIndentTitle{Vehicle Modeling Support} \cite{chronoVehicle2019}. Chrono::Vehicle is a specialized module within Chrono that offers a collection of templates (parameterized models) for various topologies of both wheeled and tracked vehicle subsystems. It also provides support for vehicle operation on rigid, flexible, and granular terrain, features closed-loop and interactive driver models, and enables both real-time and off-line visualization of simulation results. Chrono::Vehicle leverages and works in tandem with other Chrono modules, such as Chrono::FEA (for finite element support); Chrono::DEME (for granular dynamics support); Chrono::VSG, Chrono::Irrlicht, and Chrono::OpenGL (for run-time visualization); and Chrono::Multicore for parallel computing support. Chrono::Vehicle works with several terrain models -- SCM, CRM, and DEM, see below.

\begin{figure}[h]
	\centering
	\includegraphics[width=\linewidth]{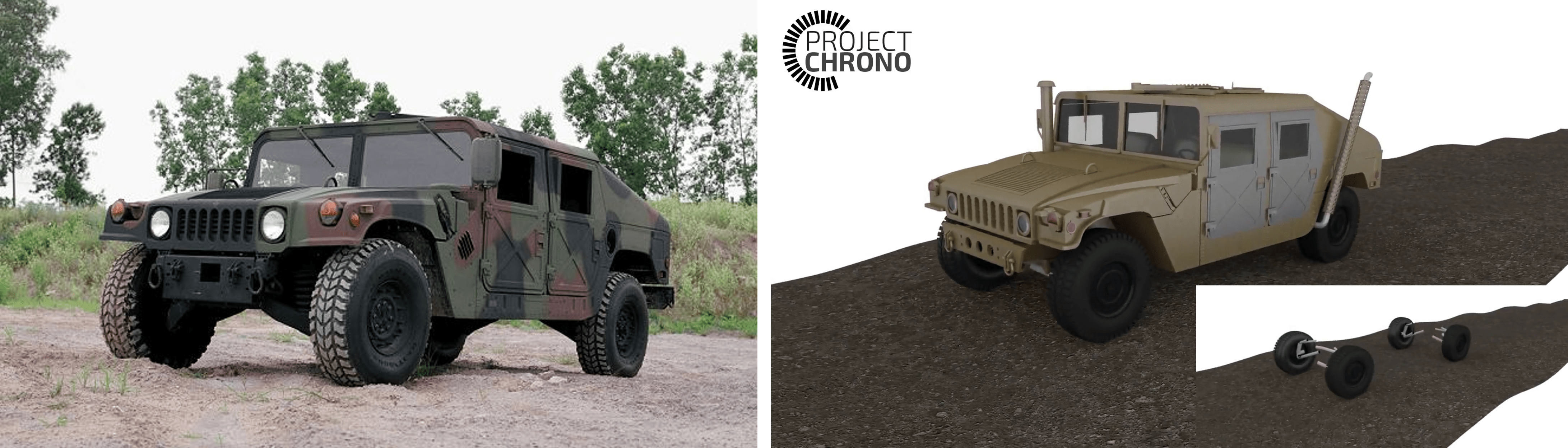}
	\caption{An example of a HMMWV vehicle and its Chrono::Vehicle digital twin.}
	\label{fig:chronoVehicle}
\end{figure}

Chrono::Vehicle provides a comprehensive set of vehicle subsystem templates (for tires, suspensions, steering mechanisms, drivelines, sprockets, track shoes, etc.), templates for external systems (for powertrains, drivers, terrain models), and additional utility classes and functions for vehicle visualization, monitoring, and collection of simulation results.  As a middleware library, Chrono::Vehicle requires the user to provide C++ classes for a concrete instantiation of a particular template.  An optional Chrono library provides complete sets of such concrete C++ classes for several ground vehicles, both wheeled and tracked, which can serve as examples for other more customized vehicle models. An alternative mechanism for defining concrete instantiation of vehicle system and subsystem templates is based on input specification files in the JSON format.  For additional flexibility and to allow integration of third-party software, Chrono::Vehicle is designed to permit either monolithic simulations or co-simulation where the vehicle, powertrain, tires, driver, and terrain/soil can be simulated independently and simultaneously.

Chrono::Vehicle currently supports three different classes of tire models: rigid, semi-empirical, and finite element. Rigid tires can be modeled as cylindrical shapes or else as non-deformable triangular meshes. From the second class of tire models, Chrono::Vehicle provides templated implementations for Pacejka (89 and 2002) \cite{pacejka2005}, Fiala \cite{TR2015-13mikeFiala}, and TMeasy \cite{Rill15} tire models, all suitable for maneuvers on rigid terrain. Finally, the third class of tire models offered are full finite element representations of the tire.  While these models have the potential to be the most accurate due to their detailed physical representation of the tire, they are also the most computationally expensive among the tire models currently available in Chrono::Vehicle. Using ANCF or Reissner shell elements, these FEA-based tire models can account for simultaneous deformation in tire and soil, for high-fidelity off-road simulations.

Tracked vehicles in Chrono::Vehicle are fully modeled as multibody systems. Templates for both segmented and continuous-band tracks are available, the latter providing options for modeling using 6-DOF bushing elements or else FEA shell elements.  Frictional contact interaction, both internal (between vehicle components) and with the terrain, relies on the underlying Chrono capabilities and supports both non-smooth (i.e., complementarity-based) and smooth (i.e., penalty-based) contact formulations.

\sbelNoIndentTitle{Sensor Simulation Support} \cite{asherSensorSimulation2021}.  Chrono::Sensor is a real-time capable sensor simulation module that provides a variety of sensors, including cameras, LiDARs, SPADs and GPS/IMU to support autonomy/Robotic simulations. At its core, Chrono::Sensor employs a ray tracing engine that utilizes the NVIDIA OptiX framework \cite{optixNVIDIA}. It uses path tracing with global illumination and PBR to simulate the interaction of light with the environment. It also has a volumetric rendering pipeline to model volumes such as fog and dust and also a transient rendering pipeline to model high fidelity Time-of-Flight sensors such as LiDARs and SPADs. On top of the standard rendering pipeline, Chrono::Sensor provides realistic sensor simulations by modeling common artifacts such as lens distortion, depth-of-field, exposure, sensor noise, sensor lag, etc. Furthermore, it was designed to work in tandem with the Chrono dynamics engine to provide real-time sensor data for the vehicle dynamics simulation. 

\begin{figure}[h]
	\centering
	\includegraphics[width=\linewidth]{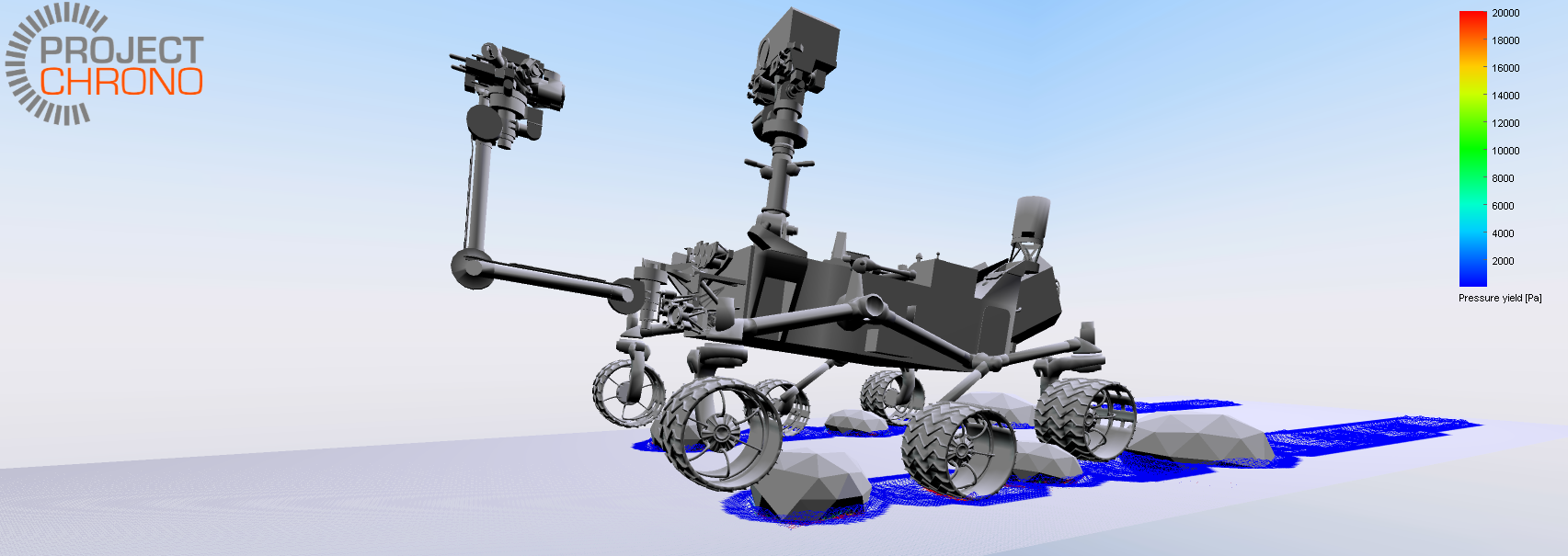}
	\caption{A Curiosity rover Chrono replica operating on SCM terrain while traversing short wavelength obstacles.}
	\label{fig:curiositySCMterrain}
\end{figure}

\sbelNoIndentTitle{Terramechanics: Soil Contact Model (SCM)}~\cite{chronoSCM_JCND_2023}. This modeling approach originates in the work reported in~\cite{Krenn2008SCM,Krenn2011}. SCM is a general-purpose model of deformable terrain that runs in real-time on commodity hardware. It is a generalization of the Bekker formula $p = \left( \frac{K_c}{b} + K_\phi \right) z^n$,  which relates the normal pressure $p$ to the sinkage $z$ for a wheel of width $b$ using a semi-empirical, experiment-based curve fitting via parameters $K_c$, $K_\phi$, and $n$~\cite{bekker56}. The pressure formula is augmented with the Janosi-Hanamoto equation~\cite{janosi61}, in which the shear stress is computed using $\tau =  \tau_\text{max} \left( 1 - e^{-j/k}\right)$, where $\tau_\text{max} = c + p \tan(\phi)$ is the maximum share stress, $j$ the accumulated shear, $c$ the cohesion, $\phi$ the internal friction angle, and $k$ the so-called Janosi parameter. Chrono's SCM implementation provides a high-performance, OpenMP-enabled \cite{openMP}, real-time capable solution~\cite{chronoSCM_JCND_2023}. Defining the real-time factor of a simulation as the amount of compute time necessary to spend to advance the state of the dynamics system forward in time by 1 second, SCM in Chrono runs at RTF of 1.0 and below for basic rover simulations on deformable terrains that do not contain short wavelength features. When the scenario simulated calls for the vehicle to cross over movable rocks that lay around on a deformable terrain, the RTF factors can go as high as 30 to 40, i.e., the simulation stops being real time. As a rule of thumb, the RTF depends on the hardware used to run the simulation, the complexity of the scenario, e.g., vehicle interacts or not with rocks, and the geometric complexity of the grousers present on the wheels. On the upside, the SCM implementation in Chrono provides a good compromise between simulation speed and accuracy. The SCM results are satisfactory under three main assumptions: the wheel sinkage is small, slip ratio is low, and the wheel geometry is close to a cylinder without lugs or grousers \cite{meiriongriffith11,Senatore2012}. A visualization mesh can be generated from the underlying SCM virtual grid to allow real time visualization of the deformed terrain with any of the Chrono run-time visualization modules or with Chrono::Sensor. This enables the placement of virtual camera sensors on autonomous platforms, thereby facilitating sinkage estimation for more robust trafficability control policies. A snapshot of a Curiosity simulation on SCM terrain is provided in Fig.~\ref{fig:curiositySCMterrain}.

\begin{figure}[!t]
	\centering
	\includegraphics[width=\linewidth]{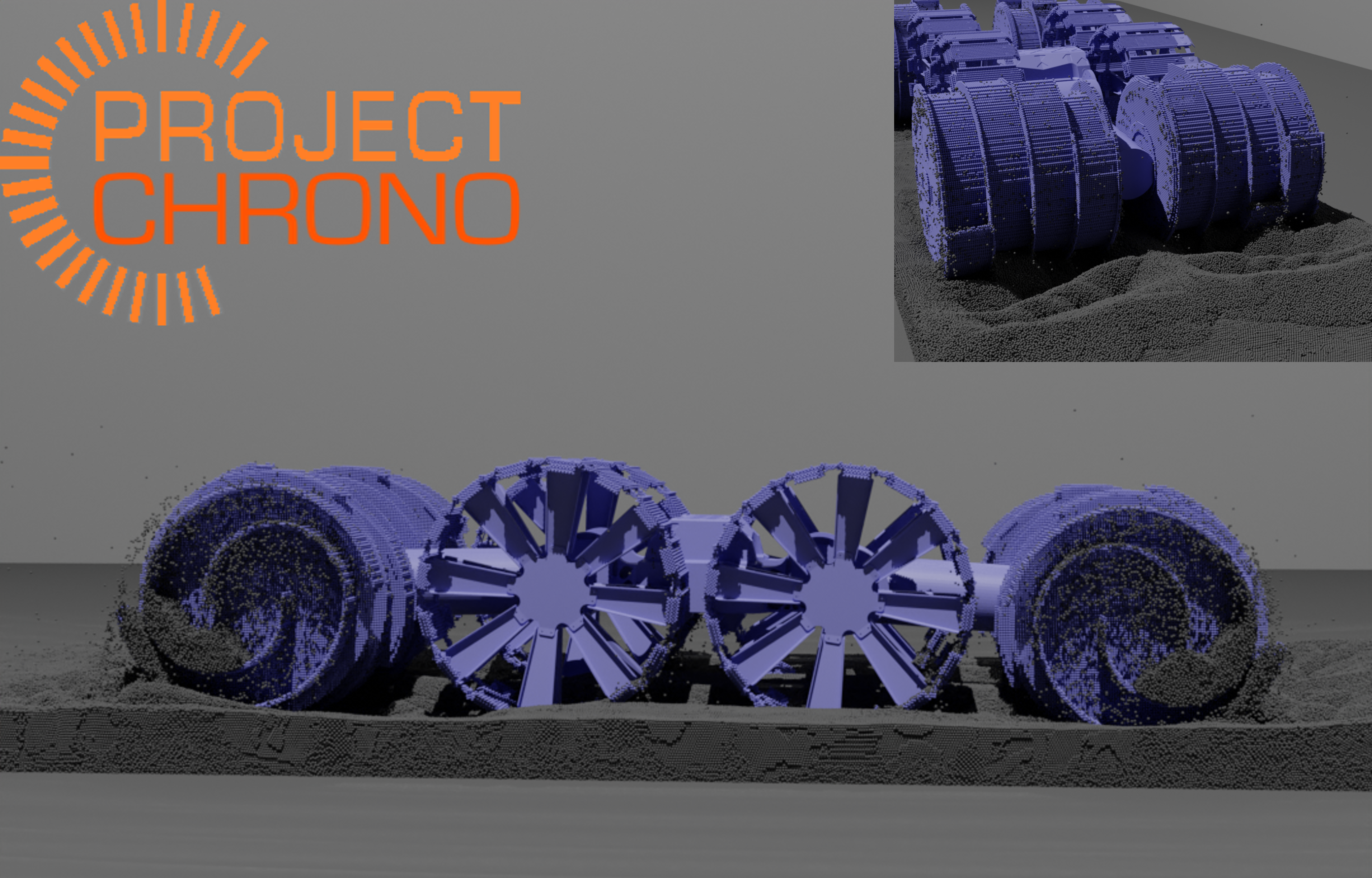}
	\caption{RASSOR excavator simulated in Chrono and rendered \textit{off-line} in Blender.}
	\label{fig:rassorLuning}
\end{figure}

\sbelNoIndentTitle{Terramechanics: Continuum Representation Model (CRM)}~\cite{weiGranularSPH2021,weiCRM2024,weiVirtualBevameter2024}. In Chrono's CRM, the terrain, even if granular, is approximated as a continuum that has a suitable chosen constitutive equation to yield a terramechanics model that trades off speed for accuracy \cite{weiGranularSPH2021,weiTracCtrl2022,weiVirtualBevameter2024}. The terrain is considered incompressible, and the field unknowns are the velocity $\mathbf{u}$ and Cauchy stress tensor $\boldsymbol{\sigma}$. The latter two variables enter the mass and momentum balance equations as in 
\begin{subequations} 
	\label{subequ:continuumBalance}
	\begin{align}
		\frac{d\rho}{d t} & =-\rho \nabla \cdot \mathbf{u} \label{subeq:MassBalance} \\
		\frac{d \mathbf{u}}{d t} & = \frac{\nabla \cdot \boldsymbol{\sigma}}{\rho}  + \mathbf{f}_b  \label{subeq:MomentumBalance} \; ,
	\end{align}
where $\rho$ is the density of the continuum and $\mathbf{f}_b$ denotes the external force per unit mass. When applied in terramechanics, for closure, a relation between the Jaumann $ \overset{\triangle}{\boldsymbol{\sigma}} $ stress rate tensor and strain tensor \cite{Monaghan2000,Gray2001sph,yue2015continuum,KamrinFluidMechanics2015} poses the stress rate tensor as 
\begin{equation}
	\dot{\boldsymbol{\sigma}} = \dot{\boldsymbol{\phi }}\cdot{\boldsymbol{\sigma}}-{\boldsymbol{\sigma}}\cdot\dot{\boldsymbol{\phi }} + \overset{\triangle}{\boldsymbol{\sigma}} \; ,
\end{equation}
\end{subequations}
where the rotation rate tensor is defined as $\dot{\boldsymbol{\phi}} =\frac{1}{2}(\nabla\mathbf{u} - \nabla{{\mathbf{u}}^{{\intercal}}})$. For a viscous incompressible fluid, Eq.~(\ref{subeq:MassBalance}) is restated as $\nabla \cdot \mathbf {u}=0$, and Eq.~(\ref{subeq:MomentumBalance}) assumes a simpler form, $ \rho \frac{d \mathbf{u}}{dt} = -\nabla p + \mu \nabla^2 \mathbf{u} + \rho {\mathbf{f}}_b $, where $ \mu $ is the dynamic viscosity coefficient and $ p $ is pressure pressure defined via $\boldsymbol{\sigma} \equiv -p{\bf I}+\boldsymbol{\tau}$,  with $\boldsymbol{\tau}$ being the deviatoric component of the Cauchy stress tensor. The set of time-dependent partial differential equations in Eq.~(\ref{subequ:continuumBalance}) are numerically solved via a spatial discretization using the smoothed particle hydrodynamics (SPH) method \cite{bui2008lagrangian,chenSPH3DgranMat2012,nguyenSPHgranFlows2017,hurley2017continuum,xu2019analysis,chen2020gpu,weiGranularSPH2021}. A snapshot of a RASSOR simulation in CRM terrain is provided in Fig.~\ref{fig:rassorLuning}. Note that the image was generated in Blender, and as a post-processing step. The camera sensor model implemented in Chrono is discussed later in this document. While the picture in Fig.~\ref{fig:rassorLuning} is eye-pleasing, it is not photo-realistic, i.e., not indicative of what an actual camera sensor, which has relatively low resolution, would typically register on the Moon in challenging light conditions. This aspect is elaborated upon in section \ref{sec:cameraSim}.

\begin{figure}[h]
	\centering
	\includegraphics[width=\linewidth]{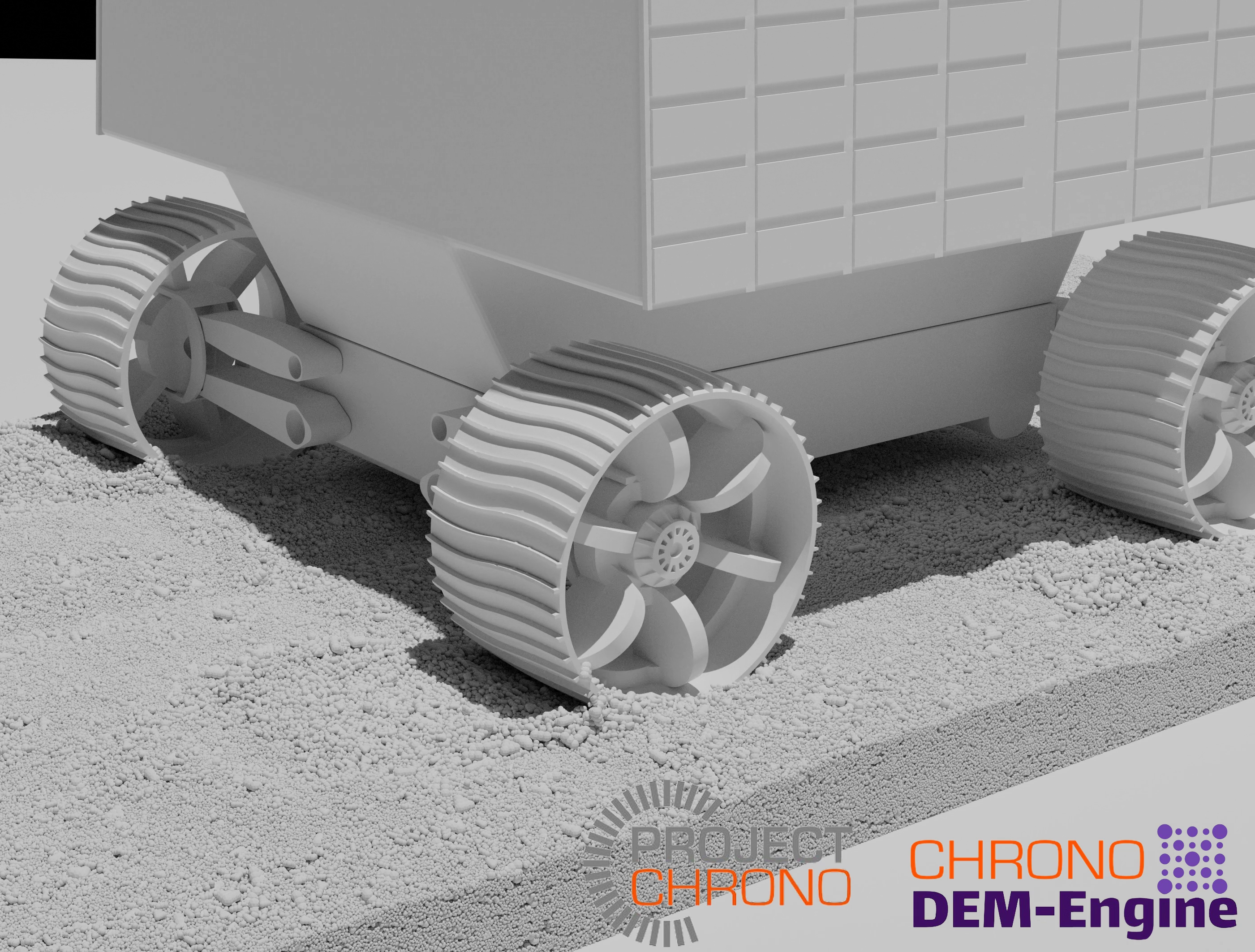}
	\caption{A VIPER Chrono replica operating on DEM terrain on a 20 degree slope.}
	\label{fig:viperDEM}
\end{figure}

\sbelNoIndentTitle{Terramechanics: Discrete Element Method (DEM)}~\cite{conlainBillion2019,ruochunGRC-DEM2023,ruochun2024dem}. Unlike CRM where the terrain is homogenized, DEM keeps track of the motion of each particle (element) in the terrain to yield a highly accurate but exceedingly expensive L1-grade terramechanics model. The equations of motion of an element $i$, for the translational degrees of freedom, are $m_i {\dot {\mathbf{v}}}_i = m_i \mathbf{g} + \sum_{j} (\mathbf{F}_n^{ij} + \mathbf{F}_t^{ij} ) $, while they are $I_i {\dot {\boldsymbol{\omega}}}_i = \sum_{j} ( \mathbf{r}^{ij} \times \mathbf{F}_t^{ij}  ) $  for the rotational degrees of freedom. Here $m_i$ and $I_i$ are the mass and mass moment of inertia of particle $i$; $j$ indexes particles in contact with $i$. The normal and tangential friction forces, $\mathbf{F}_n^{ij}$ and $\mathbf{F}_t^{ij}$, assume the form of a spring-damper mechanism, $\mathbf{F}_n = k_n \mathbf{u}_n + \gamma_n \mathbf{v}_n $ and $\mathbf{F}_t = k_t \mathbf{u}_t + \gamma_t \mathbf{v}_t$, where the stiffness and damping coefficients, $k_n$, $k_t$, $\gamma_n$, and $\gamma_t$, are derived from material properties, particle shape and effective mass; the normal penetration $\mathbf{u}_n$ is produced from collision detection; the tangential displacement $\mathbf{u}_t$ is based on contact history \cite{jonJCND2015}; and the relative velocity at the contact point, $\mathbf{v}_n$ and $\mathbf{v}_t$, are derived from contact kinematics. Coulomb friction condition can be imposed by capping the tangential force, $\|\mathbf{F}_t\| \leq \mu \|\mathbf{F}_n\|$, with $\mu$ being the friction coefficient. For complex DEM simulations of mono-dispersed particles, Chrono has scaled to billions of degrees of freedom on one GPU card \cite{conlainNicDan2020}. A frame of a VIPER simulation on DEM is shown in Fig.~\ref{fig:viperDEM}.

\begin{figure}[h]
	\centering
	\includegraphics[width=\linewidth]{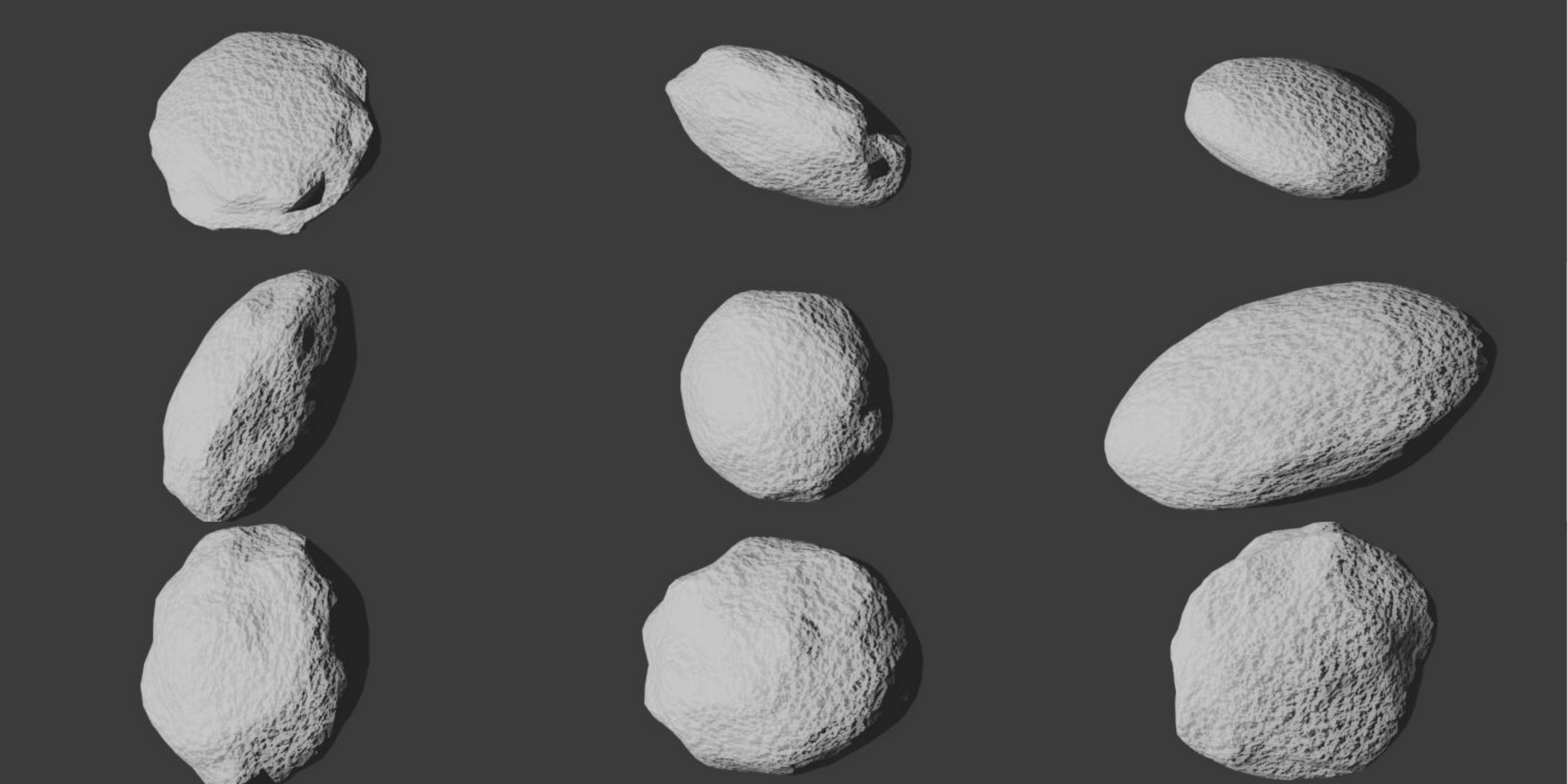}
	\caption{A sample of regolith shapes used for visualization of the CRM terrain}
	\label{lunar_granular}
\end{figure}

\sbelNoIndentTitle{Sensing Deformable CRM Terrains.} Being able to simulate high fidelity deformable terrain mechanics allows us to simulate complex wheel-terrain interactions between the rovers and lunar regolith. One such example, is modeling the wheel sinkage and slip of a rover. Our simulator, provides the capability, to visualize these wheel-terrain interactions from the perspective of an onboard wheel camera such that, those synthetic data can be used to train Visual Wheel Sinkage Estimation Algorithms (VWSE). The challenge with visualizing CRM simulations, is that they have a mesh-free particle representation of the terrain. As such, we opted to use voxel ray tracing rendering methods to render the terrain through our sensor simulation module. We leverage, the NanoVBD~\cite{Museth2021NanoVDB} library to create a voxel grid on the GPU from the terrain particles. Then, we use voxel rendering to render each voxel as a user defined geometry. In order to achieve visual realism in rendering of lunar regolith, we established a pipeline to generate random meshes of granular lunar regolith particles, which we used to substituted the spherical SPH particles. A collection of such randomly generated granular meshes is shown in Figure \ref{lunar_granular}. We then randomly associate each of these meshes to a voxel in the voxel grid. Our pipeline is able to render a simulation with approximately 8-10M particles at 20 fps on a single GPU. 

\sbelNoIndentTitle{Dust Simulation.} In lunar or other environments with low gravitational pull and a rarefied atmosphere, dust poses several challenges, one of which is its adverse impact on the perception process. Dust can obstruct the view of cameras and adversely impact active light sensing technologies, complicating navigation and task execution. For instance, when estimating the sinkage of a rover's wheels, dust can obscure the wheel-terrain interaction, hindering accurate measurements of how deeply the wheels are sinking. Chrono has a basic phenomenological model for dust production and propagation, and implements a volumetric rendering-based pipeline to render dust volumes within Chrono::Sensor. For a physics-based alternative, the user is referred to \cite{plumeSim2024}. The highlights of the model are as follows:
\begin{itemize}
	\item The dust particles represent a collection of points associated with a point cloud. 
	\item In the absence of an atmosphere, the only force acting on a dust particle is the lunar gravitational pull. Consequently, once the particle is set in motion, it behaves as a material point under the influence of gravity. Since this type of motion has a closed-form solution, there is no need for numerical integration methods.
	\item The motion of the dust particles is tracked on the GPU. 
	\item Each CRM marker emits dust particles if ($i$) it's on the surface of the terrain, and ($ii$) the CRM marker's speed is higher than a threshold value. Static CRM markers do not emit dust particles.
	\item The ``dust emission \& propagation'' algorithm is executed as a post-processing step and it happens online but at intervals that are independent of the time step used to compute the dynamics of the terrain. 
	\item Currently, dust particles do not collide with each other. In other words, if they are on a collision course, they simply ``tunnel'' through one another and continue moving along their original trajectories.	 
	\item Once a dust particle reaches the soil it is terminated.
	% \item Once a sphere hits a surface that is not the soil, it sticks to it. The net effect of ``sticking to it'' is that the particle contributes to changing the texture of the surface hit by the dust particle. Additionally, the dust particle vanishes from the simulation just as it vanishes when hitting the ground.
\end{itemize}
In terms of limitations, the current implementation only works with CRM terrains and does not account for electrostatic interactions, such as the attraction of particles due to surface charge or polarization effects. Dust settles on a surface solely due to physical collision, without any attraction from the surface itself.
%  and the net effect of this process is a change of the surface's texture.

\begin{figure}[h]
	\centering
	\includegraphics[width=0.45\textwidth]{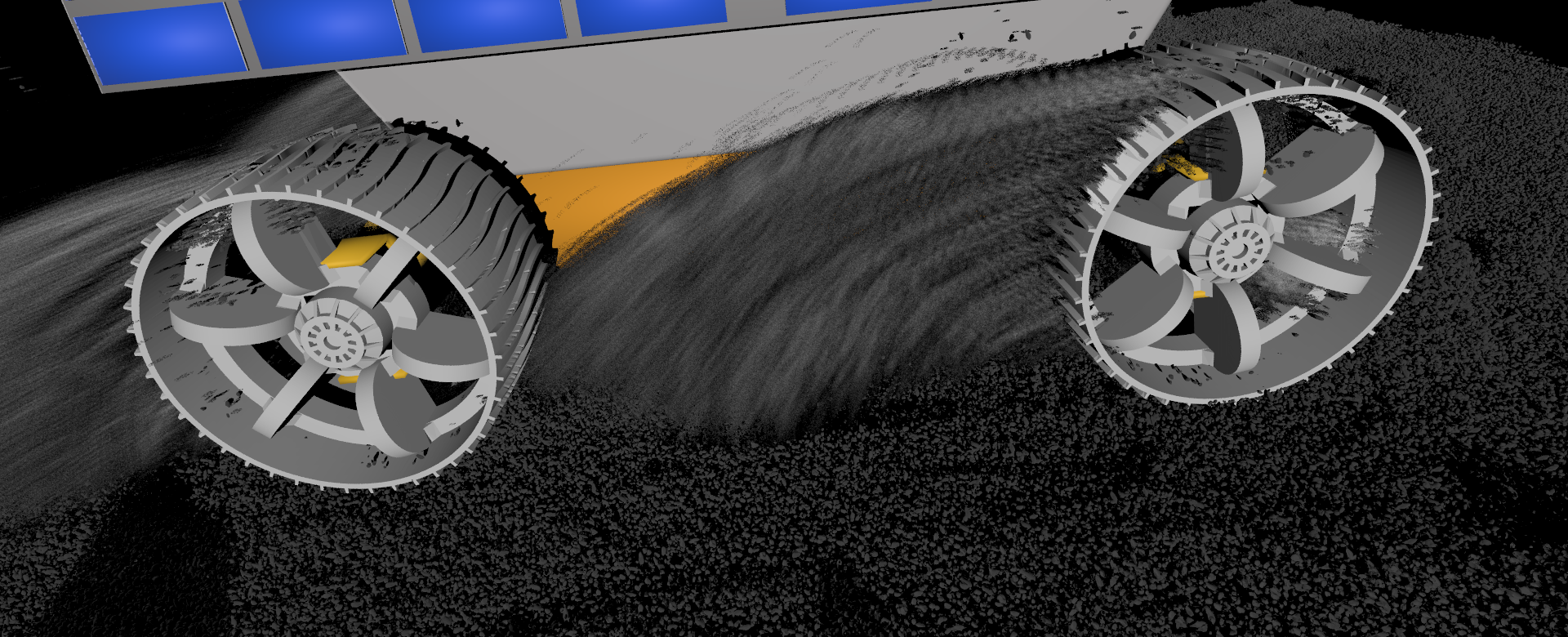}  % 
	\includegraphics[width=0.45\textwidth]{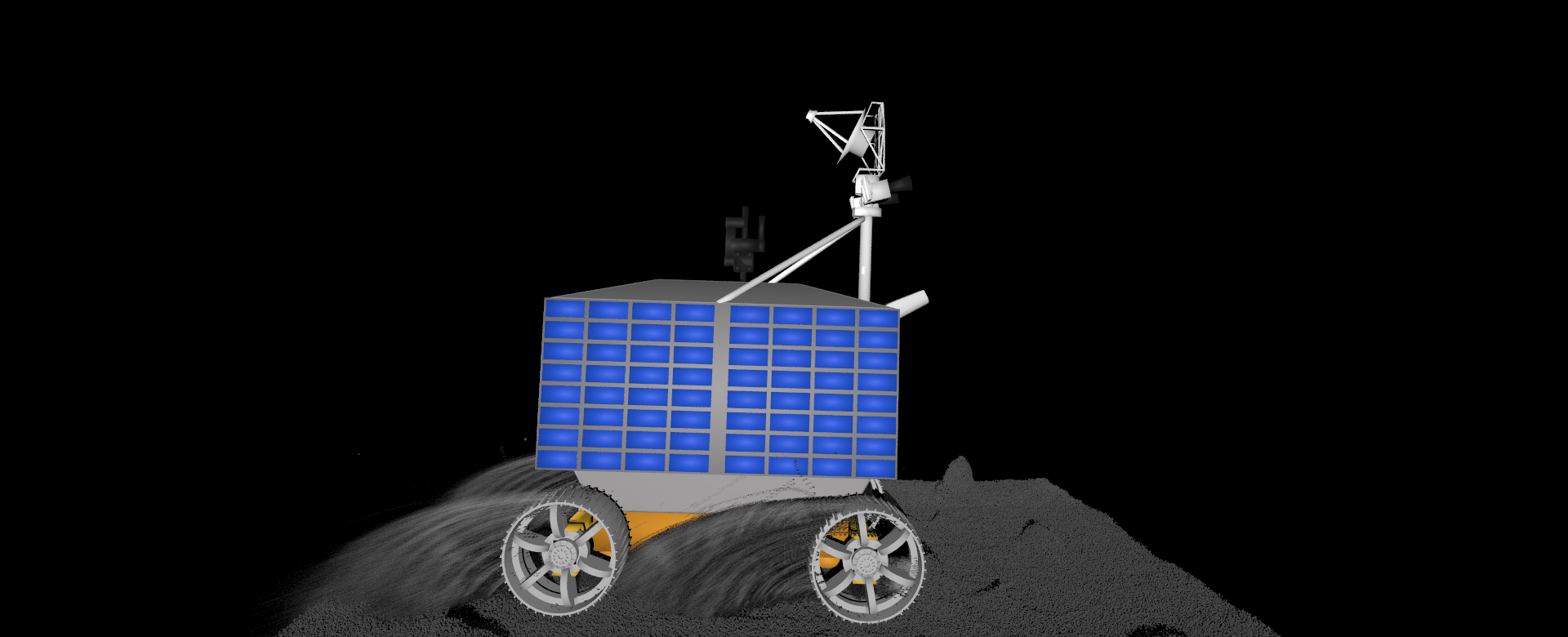}  % 
	\caption{View from the wheel camera (top) and a third-person-view (bottom) of VIPER rover traversing lunar regolith and producing a dust trail.}
	\label{fig:dust_rendering}
\end{figure}

We visualize the dust particles using a volumetric rendering pipeline. The dust cloud is represented as a NanoVDB density grid \cite{Museth2021NanoVDB}, where each voxel contains the density of the dust cloud at that location. We then use a ray marching algorithm to render the dust cloud using a volumetric rendering method \cite{FongVolumeRendering2017}, which models the scattering and absorption of light as it passes through the dust cloud. The result is a realistic representation of the dust cloud that can be visualized in our simulator. We model the dust volume as a high absorption, low scattering volume, but this is was chosen based on visuals and not based on any study on the actual scattering and absorption properties of lunar dust. A more physically grounded paramterization of the volumetric properties of lunar dust and how it compares to the Hapke parameters and lunar regolith properties is part of our future work in this direction. Figure \ref{fig:dust_rendering} shows a dust trail produced by the VIPER rover traversing a lunar terrain modeled using CRM.

% \begin{figure}[t!]
% 	\centering
% 	\includegraphics[width=1.0\columnwidth]{images/viper_dust.pdf}
% 	\caption{View from the wheel camera (top) and a Third-person-view (bottom) of Viper rover traversing lunar regolith and producing a dust trail.}
% 	\label{dust_rendering}
% \end{figure}

\sbelNoIndentTitle{Chrono::ROS}.  To facilitate software (autonomy stack related) and hardware (chip, that is) in the loop design and testing, the Chrono::ROS module accommodates ROS2-based \cite{ROS2,ROS-2009} robotics algorithms within Chrono simulations. By enabling the ROS2 communication network and common functionalities -- such as publishers, subscribers, and topics, users can design and deploy their autonomy solutions more efficiently using Chrono-simulated agents. Chrono::ROS serves as a communication bridge between the simulation and the autonomy stack: it can output standard ROS2 topics or services that expose simulated sensor signals, vehicle state information, and similar quantities relevant in the context of autonomy stack design. The Chrono::ROS module supports handlers that convert simulated information (e.g., Chrono::Sensor messages or Chrono::Vehicle states) to ROS2 topics through simple API function calls. Users can also implement their own custom handlers to facilitate communication between simulation agents and autonomy algorithms.

\sbelNoIndentTitle{Chrono: Miscellaneous other components}. This section has covered Chrono components that come into play in lunar ground operations and terramechanics applications. This paragraph contains a concise summary of other Chrono components that are less relevant in the context of extraterrestrial applications. Chrono::Engine provides multibody-dynamics and nonlinear finite element analysis core functionality, with robust treatment of friction and contact that draws on two approaches: penalty method and complementarity-based method. Chrono::SolidWorks  \cite{chrono-solidworks} provides interoperability with SolidWorks, e.g., the ability to import into Chrono mechanical systems defined in SolidWorks. The Chrono::Vehicle module also includes a hybrid parallel (MPI-OpenMP-CUDA) co-simulation framework for vehicle-terrain interaction, using any of the available Chrono terrain models or a third party terramechanics solver \cite{serbanCosimIJVP2019}. In this setup, for instance, four processes would run the tire models on a vehicle, one process would run the chassis dynamics, and one process would run the terrain simulation. This reduces the simulation time by using six different cores to run one simulation. Chrono::FSI provides support for computational fluid dynamics and is essentially a partial differential equations (PDEs) solver that does spatial discretization using the Smoothed Particle Hydrodynamics (SPH) method \cite{liu2003smoothed}. This SPH solver is also used in terramechanics for CRM terramechanics. The Chrono::DEME module provides support for carrying out Discrete Element Method simulations on the GPU. Irrlicht \cite{irrlicht}, Vulkan Scene Graph \cite{vulkan}, and OpenGL \cite{opengl} are used for run-time visualization; Chrono::Unity \cite{chrono-unity} provides integration with the Unity Game Engine \cite{unityGaming}. Chrono::MATLAB provides interoperability with MATLAB and Simulink, while Chrono::FMI provides support for the functional mock-up interface that enables co-simulation support \cite{FMI}. Chrono::PardisoMKL provides an interface for Intel's Math Kernel Library in general, and the Pardiso direct solver in particular. SynChrono \cite{synchrono2020} is a framework that enables the time and space coherent simulation of large collections of agents, e.g., rovers, vehicles. The time coherence enforced by SynChrono ensures that all vehicles share the same global time, which avoids one vehicle rushing into the future while other lagging into the past -- they advance their state on the same heartbeat. The space coherence enforced by SynChrono enables the vehicles to sense each other as well as the way they change the environment in which they operate (for instance, Vehicle A leaves ruts behind in the terrain, which should be picked up by Vehicle B even though Vehicle B is run by a different Chrono process). PyChrono \cite{pychrono2022} is a SWIG-based wrapper of the C++ Chrono library that allows one to import this module and run Chrono simulations from Python. Gym Chrono is a set of simulated environments for deep Reinforcement Learning extending OpenAI Gym \cite{openAI-gym2016} with robotics and autonomous driving tasks, to which end it draws on PyChrono. 

%% file: sections/camerasim.tex
%{\sbelComment{Input from Bo-Hsun.}}

\begin{figure*}[!t]
	\centering
	\includegraphics[width=2.0\columnwidth]{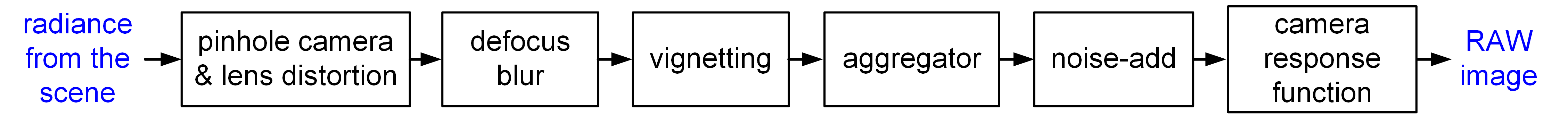}
	\caption{The framework of the physics-based virtual camera sensor.}
	\label{fig:phys_cam_struct}
\end{figure*}

Although several state-of-the-art camera simulators based on physically-based rendering (PBR) exist, they either take an extended amount of time to synthesize a single image \cite{lyu2022validation}, making them difficult to incorporate into real-time robotics simulations, or lack complete control over essential camera settings (e.g., ISO or focus distance) for users \cite{li2023behavior-1k}. In this section, we discuss a physics-based camera model that offers comprehensive camera settings to control and includes a broad range of optical artifacts \cite{asherSensors2020}. Furthermore, each model layer creatively introduces \textit{model gain parameters} that allow users to calibrate the virtual camera based on their real cameras, effectively grounding the virtual simulations in real-world conditions. Compared to conventional camera simulators, which offer limited controllability and customization, our model exposes a wide range of camera settings and parameters, making it suitable for specialized applications in extraterrestrial exploration.

The framework of the proposed physics-based camera in Chrono::Sensor is depicted in Fig.~\ref{fig:phys_cam_struct}. It extends the existing ray-tracing pinhole camera with lens distortion at its core in Chrono::Sensor \cite{asherSensorSimulation2021}, by incrementally incorporating image processing algorithms to emulate various optical artifacts. Each processing stage leverages CUDA programming to concurrently handle pixel-wise data, enabling high-performance computation. The architecture provides control over several camera settings, including the F-number, exposure time, ISO, focus distance, and focal length. Adjusting these parameters allows users to manipulate the resulting optical artifacts. The following paragraphs detail the physical principles and mathematical models applied at each stage.

\sbelNoIndentTitle{Pinhole Camera Model and Lens Distortion.} In the initial stage, every pixel data is generated using the ideal pinhole camera model and ray-tracing. The camera's location establishes the origin of the ray generator, with rays distributed uniformly across the field of view (FOV) through all pixels, as determined by the equation:
\begin{equation}
	\label{eq:fov}
	FOV = 2 \tan^{-1}\left( \frac{w}{2f} \right) \; ,
\end{equation}
where $w$ represents the effective width of the image sensor and $f$ denotes the focal length. After emission, rays are deflected according to the lens distortion model, interact with scene objects, and ultimately reach the light source. The output at this stage consists of the RGB components' irradiance, measured in $[W/m^2]$.

\begin{figure}
	\centering
	\includegraphics[width=0.8\columnwidth]{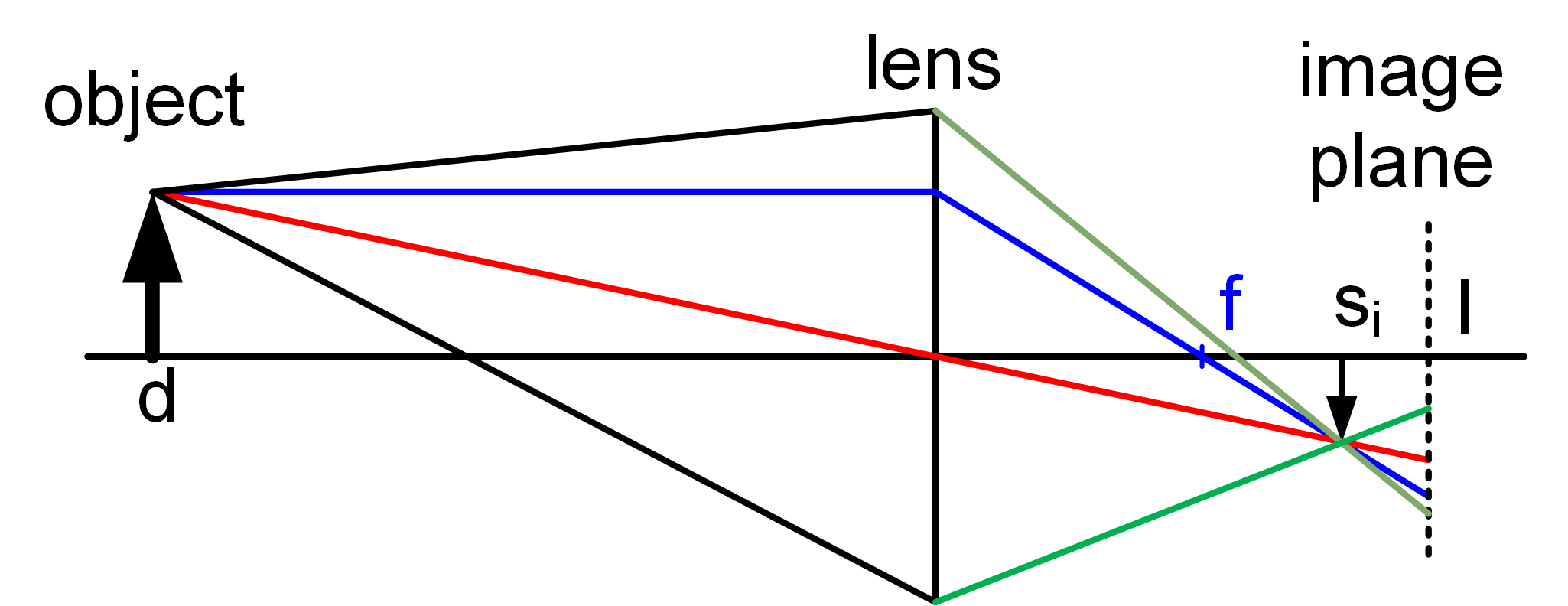}
	\caption{Physical depiction of defocus blur, based on the thin lens model.}
	\label{fig:defocus_blur_phys}
\end{figure}

\sbelNoIndentTitle{Defocus Blur.} The second stage introduces defocus blur through image processing to simulate depth of field. Using the Gaussian thin lens model, the defocus blur diameter is determined as shown in Fig.~\ref{fig:defocus_blur_phys}. With the camera's image plane distance set, the Gaussian thin lens equation fixes the scene's focal plane. Points in the scene not lying on this focal plane appear as diffused circles on the camera's image plane. The diameter of these circles of confusion, referred to as the \textit{defocus-blur diameter}, is computed in units of pixels using the formula:
\begin{subequations}
	\begin{equation}
	\label{eq:defocus_blur}
	D_{ij} = \frac{G_{defocus} \cdot f^2 \cdot \left| d_{ij} - U \right|}{N \cdot C \cdot d_{ij} \cdot (U - f)} \;,
	\end{equation}
where $G_{defocus}$ is the user-defined defocus gain, $f$ is the focal length, $N$ is the F-number, $C$ is the pixel size, $U$ is the focus distance, and $d_{ij}$ is the distance from the camera to the pixel's scene location, with $(i, j)$ marking the pixel index. Defocus blur is practically implemented by applying a Gaussian blur kernel, whose size varies per pixel:
	\begin{equation}
		\label{eq:defocus_blur_program}
		y_{ij} = \sum \limits_{k=i - \frac{D_{ij}}{2}}^{i + \frac{D_{ij}}{2}} \sum \limits_{l=j - \frac{D_{ij}}{2}}^{j + \frac{D_{ij}}{2}} \frac{x_{ij}}{2 \pi \sigma^2} \exp \left( -\frac{(k - i)^2 + (l - j)^2}{2 \sigma^2} \right)  \;,
	\end{equation}
\end{subequations}
where $y_{ij}$ is the output pixel, $x_{ij}$ is the input pixel, and $\sigma$ is the standard deviation, computed as $D_{ij}/6$. The result of this stage is still quantified in irradiance.

\sbelNoIndentTitle{Vignetting.} Vignetting is enhanced through a user-defined falloff gain, resulting in brightness reduction from the center to the periphery of the image, according to a $\cos^4\theta$ pattern, where $\theta$ is the angle between the ray from the pixel toward the aperture and the optical axis. This attenuation occurs because irradiance decreases with the projected areas of the aperture and the pixel (both diminishing with $\cos\theta$) and with the square of the distance from the aperture to the pixel (decreasing with $\cos^2\theta$). The equation is:
\begin{subequations}
	\begin{equation}
		\label{eq:vignet}
		\left( 1 - \frac{y_{ij}}{x_{ij}} \right)= G_{vignet} \cdot \left( 1 - \cos^4 \left( \theta_{ij} \right) \right) \;,
	\end{equation}
with the ray angle $\theta_{ij}$ defined as
	\begin{equation}
		\label{eq:ray_angle}
		\theta_{ij} = \tan^{-1} \left( \frac{\sqrt{a_{ij}^2+b_{ij}^2}}{f} \right) \;,
	\end{equation}
\end{subequations}
where $y_{ij}$ is the output pixel, $x_{ij}$ is the input pixel, $(a_{ij}, b_{ij})$ are the pixel's coordinates on the image plane, and $G_{vignet}$ is the vignetting falloff gain. The output at this stage is still measured in irradiance.

\sbelNoIndentTitle{Aggregator.} The aggregator integrates irradiance over time and pixel area to calculate the total electron energy per pixel. The process is defined by:
\begin{equation}
	\label{eq:aggregator_program}
	y_{ij} = G_{aggregator} \cdot x_{ij} \cdot \frac{C^2}{N^2} \cdot t \cdot QE_{R/G/B} \;,
\end{equation}
where $y_{ij}$ is the output pixel, $x_{ij}$ is the input pixel's irradiance, $G_{aggregator}$ is the user-defined sensitivity gain, $C$ is the pixel size, $N$ is the F-number, $t$ is the exposure time, and $QE_{R/G/B}$ is the quantum efficiency for each RGB channel. The output is measured in joules, indicating the total electron energy in each pixel.

\sbelNoIndentTitle{Noise Addition.} Incorporating noise is a critical aspect of achieving realistic photo simulations. The Chrono model considers photon shot noise, time-dependent noise (dark current and hot pixels), and time-independent noise (readout and fixed pattern noise (FPN)). Firstly, the output from the aggregator layer is added with the time-dependent noise and modeled as 
\begin{equation}
	\label{eq:dark_current}
	\mu = y_{aggregator} + D \cdot t \;,
\end{equation}
where $y_{aggregator}$ is the aggregator's output, $D$ represents the dark current and hot pixel rate, and $t$ is the exposure time. The full noise model is given by:
\begin{equation}
	\label{eq:noise_model}
	\begin{aligned}
		&Y = L + N \\
		&L \sim Poisson(\mu)  \approx Gaussian(\mu, G_{noise}^2 \mu) \\
		&N \sim Gaussian(0, \sigma_{read}^2) \;.
	\end{aligned}
\end{equation}
This Poisson-Gaussian noise model defines $Y$ as the random variable output from the noise-adding layer composed of two noise sources. $L$ represents photon shot noise, which is modeled as a Poisson distribution and approximated by a Gaussian distribution due to large photon counts and the Central Limit Theorem. A user-assigned gain $G_{noise}$ is creatively introduced to decide the variance of the time-dependent noises. Finally, $N$ represents the readout noise and FPN, modeled as a Gaussian distribution with the other user-assigned gain $\sigma_{read}$ multiplied to decide the scale of the time-independent noises.

\sbelNoIndentTitle{Camera Response Function (CRF).} In the final stage, the analog signal is amplified and digitized through an analog-to-digital converter. For high dynamic range (HDR) sensing under lunar conditions, a 16-bit depth format is adopted. The ISO value defines the analog amplification gain, and the CRF can be different functions based on user selection:
\begin{equation}
	\label{eq:crf}
	\begin{aligned}
		&\text{(Gamma correct)}\quad y_{ij} = a \cdot (\log_2(ISO \cdot x_{ij}))^{\gamma} + b \\
		&\text{(Sigmoid)}\quad  y_{ij} = \frac{1}{1 + e^{-a \cdot \log_2(ISO \cdot x_{ij}) - b}} \\
		&\text{(Linear)}\quad  y_{ij} = a \cdot ISO \cdot x_{ij} + b \;,
	\end{aligned}
\end{equation}
where $y_{ij}$ is the output digital value of the pixel, $x_{ij}$ is the input signal of the pixel, and $a$, $b$, and $\gamma$ are user-defined CRF parameters. For example, sigmoid functions are typical for traditional film cameras, while modern digital cameras often follow linear functions. The final outcome of this camera model is the generation of final RAW images, with pixel values in the RGB channels ranging from 0 to 65535.

\sbelNoIndentTitle{Hapke Bidirectional Reflectance Model (BRDF).} Accurately simulating lunar sensing requires proper representation of how light behaves on the Moon, primarily influenced by lunar regolith, a low-albedo, retroreflective material. The absence of atmospheric scattering and the properties of the regolith create harsh lighting, long shadows, and high contrast between illuminated and shadowed areas. The most notable phenomenon is the opposition effect, where the lunar surface appears brighter when viewed in the direction of illumination due to shadow hiding, where shadows disappear as viewing and illumination angles align \cite{kuzminykh2021physically}. Therefore, it is essential to implement a reflection model that captures these lunar characteristics.

The Hapke model, is widely used to describe the reflectance of celestial bodies with regolith. The model depicts the lunar surface as a layer of cylinders oriented toward the direction illumination, minimizing attenuation in the incident direction \cite{Hapke1963}. The modern Hapke model introduces factors like multiple scattering, surface roughness, coherent backscatter, regolith particle size, and porosity, increasing the number of parameters from five to nine \cite{Hapke2012}. For our rover simulation framework, we have implemented the modern Hapke model with some simplifications. Specifically, the Hapke Bidirectional Reflectance Distribution Function (BRDF) implemented in Chrono is expressed as,
\begin{equation*}
	\begin{split}
		f_{hapke} (i,e) = & \, \frac{K \cdot \omega \cdot LS(i_e,e_e)}{4\pi \mu_0} \\
		& \times \left[p(g)\left(1 + B_{s_0}B_s(g)\right) + M(i_e,e_e)\right] \\
		& \times \left(1 + B_{c_0}B_c(g)\right) \cdot S(i,e,\psi) \; ,
	\end{split}
\end{equation*}
where $i$ is the angle between the surface normal and the light direction, $e$ is the angle between the surface normal and the reflection direction, $g$ is the phase angle (the angle between $i$ and $e$), and $\psi$ is the azimuth angle, which is the projection of $i$ and $e$ on the ground surface. The terms $\mu_0 = \cos(i)$ and $\mu = \cos(e)$ represent the cosine of the angles of incidence and emission, respectively. In this equation, $S(i,e,\psi)$ represents the surface roughness function, $p(g)$ is the average single particle scattering phase function, $B_{s0}$ describes the opposition effect caused by shadow hiding, and $B_{c0}$ accounts for the opposition effect due to coherent backscatter (which is ignored in the implementation). Additionally, $M(i,e)$ is the isotropic multiple-scattering approximation function, and $LS(i,e)$ follows the Lommel-Seeliger law, which models isotropic single scattering in the regolith. For further details regarding these parameters and functions, the interested reader is referred to Appendix A of \cite{HapkeParams2014}.

Table~\ref{table:hapke_params} reports the values for the above parameters currently used in our model. The values were obtained from \cite{kuzminykh2021physically,HapkeParams2014}. The parameters $w$, $b$, $c$, $B_{s_0}$ and $h_s$ are functions of wavelength and vary based on the region of the lunar surface \cite{HapkeParams2014}. In order to obtain a single weighted average value, each of those functions was weighted using the luminous efficiency function such that the final weighted values are adapted to the human perception \cite{kuzminykh2021physically}. In our implementation, we ignore the effcts from coherent backscatter due its minimal impact. Note that the aforementioned parameter functions are limited to the $30^{\circ}$ S to $30^{\circ}$ N region of the Moon, which is the equatorial/central region of the Moon \cite{HapkeParams2014}.

\begin{table}[h]
	\centering
	\rowcolors{1}{white}{gray!15}  % Alternates row colors: white and light gray
	\begin{tabular}{>{\centering\arraybackslash}m{1.5cm} >{\centering\arraybackslash}m{1.5cm} m{4cm}}
		\toprule
		\textbf{Parameter}     & \textbf{Value}              & \textbf{Description} \\ \midrule
		$w$           & 0.03257            & Single-scattering albedo: Ratio of scattered energy to extinction (attenuated) energy. \\
		$b$           & 0.23955            & Shape-controlling parameter: Controls the amplitude of the forward and backward scatter of a particle. \\
		$c$           & 0.30452            & Weighting parameter: Controls contribution of backward and forward scatter at a particle. \\
		$B_{s_0}$     & 1.80238            & Amplitude of the opposition effect caused by shadow hiding. \\
		$B_{c0}$      & 0 (ignored) & Amplitude of the opposition effect caused by coherent backscatter. \\
		$h_s$         & 0.07145            & Angular width of the opposition effect caused by shadow hiding. \\
		$h_c$         & 1 (ignored) & Angular width of the opposition effect caused by coherent backscatter. \\
		$\Phi$        & 0.3                & Filling factor: Opposite to porosity. Reflectance increases as porosity decreases. \\
		$\theta_p$    & 23.4$^{\circ}$     & Photometric roughness: Controls the surface roughness. \\
		\bottomrule
	\end{tabular}
	\caption{Parameter descriptions and values for the Hapke BRDF model}
	\label{table:hapke_params}
\end{table}

% \sbelComment{TODO: Add a picture of sim hapke model}

%% file: sections/experiments.tex
The ``sensing in lunar conditions'' Chrono framework will be demonstrated in three contexts: rendering scenes in low light using the Hapke BRDF for both VIPER and RASSOR; evaluating the accuracy of stereo depth estimation; and object detection. For depth estimation, images will be passed to a third-party, data-driven depth estimator called IGEV \cite{xu2023iterative}. The key question is whether there is a bias in the size of depth estimation errors for real versus synthetic images. Finally, for the object detection task, synthetic images generated in Chrono::Sensor, along with real-world images, will be passed to YOLOv5 \cite{yolov5} to determine if, statistically speaking, the ``judge'' (i.e., YOLOv5) performs equally well at detecting objects in real images as it does when handed synthetic images.

\sbelNoIndentTitle{Case Study 1: VIPER \& RASSOR on CRM Terrain}

\begin{figure}[!t]
    \centering
    \includegraphics[width=1.0\columnwidth]{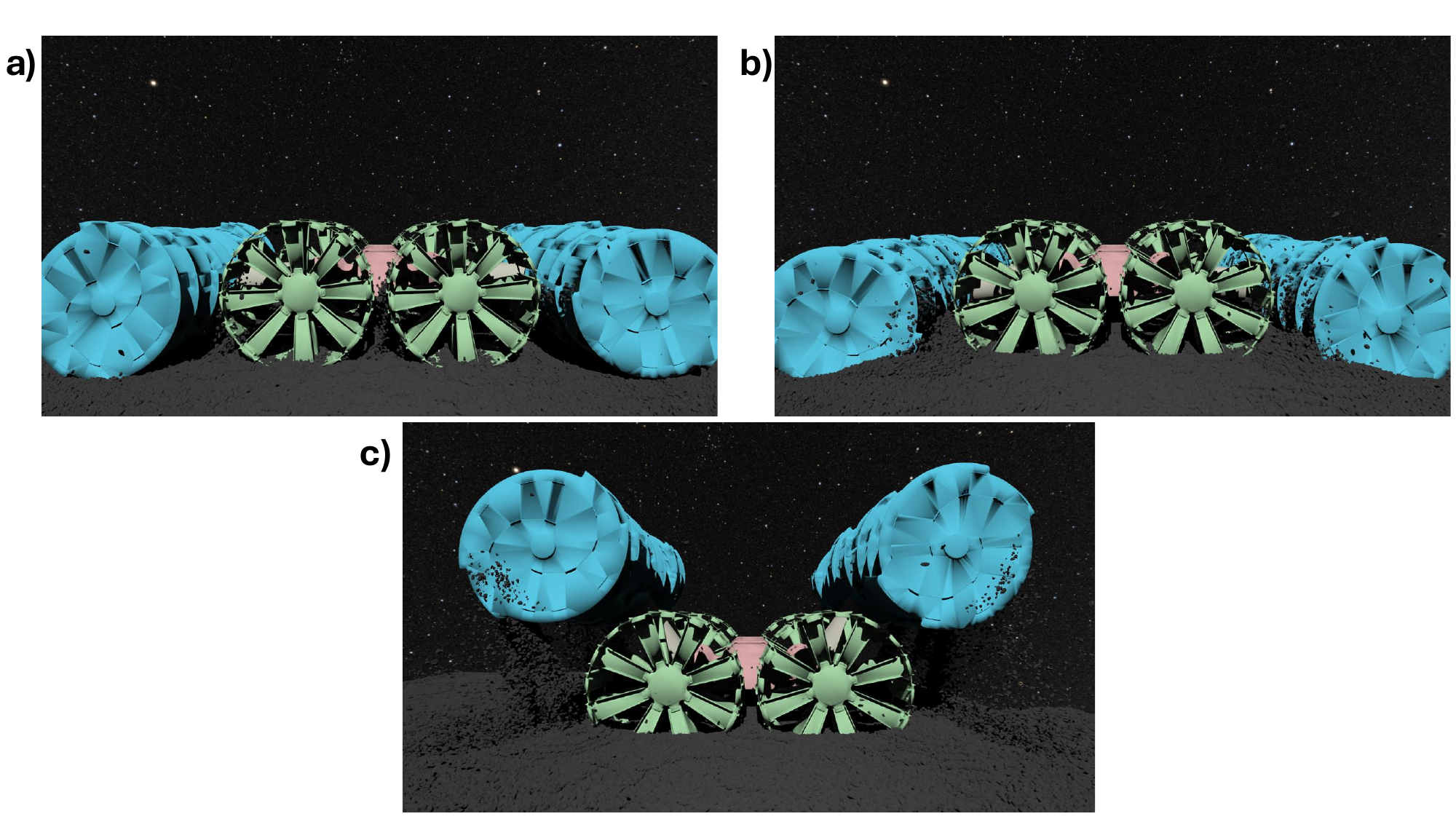}
    \caption{RASSOR rover \textbf{a)} Traversing, \textbf{b)} Excavating lunar regolith and \textbf{c)} Dumping regolith.}
    \label{fig:rassor_normal}
\end{figure}

\begin{figure}[!t]
    \centering
    \includegraphics[width=0.8\columnwidth]{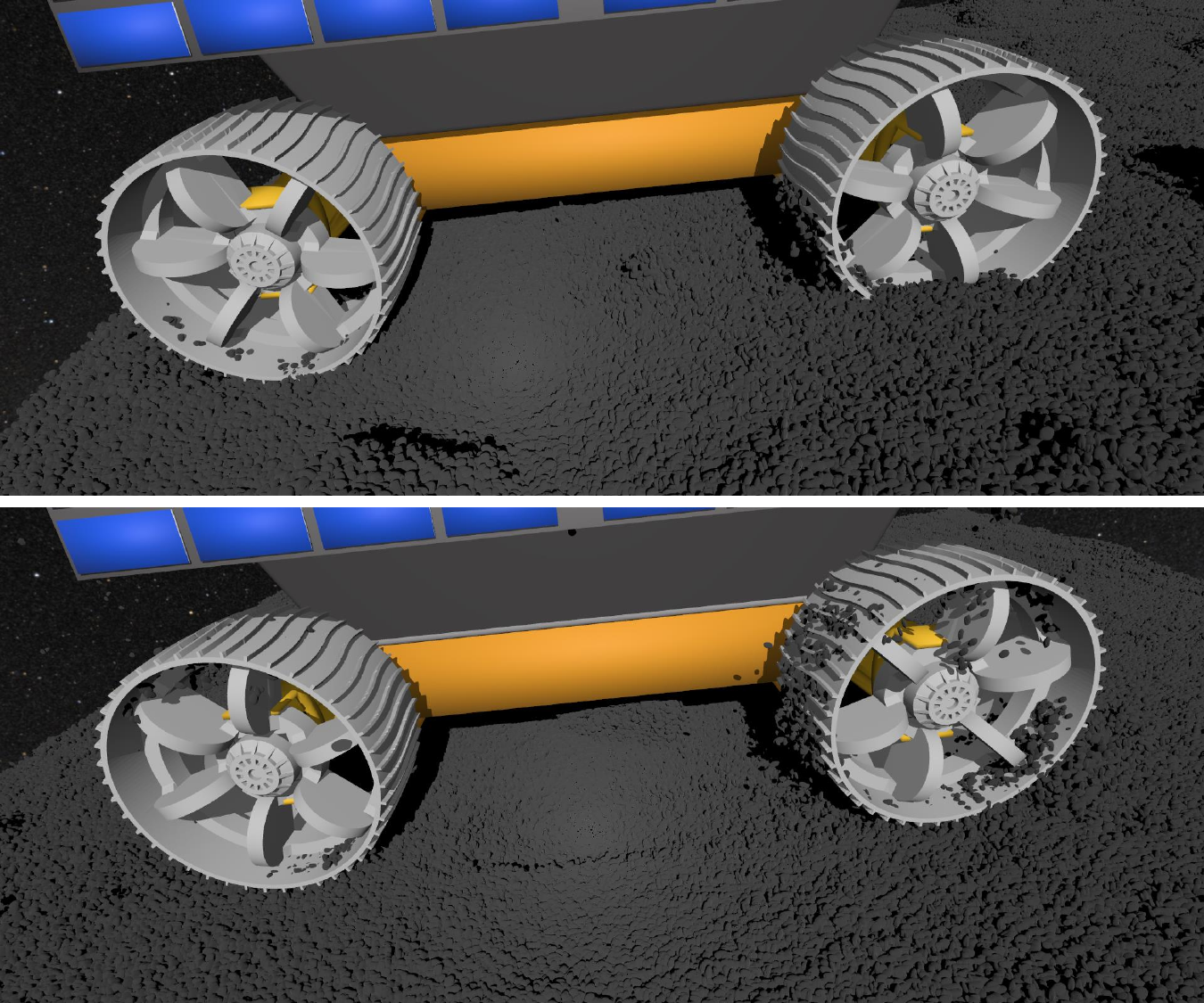}
    \caption{Viper rover traversing lunar regolith.}
    \label{fig:viper_normal}
\end{figure}

We ran simulations of the VIPER rover traversing a rough lunar terrain with rocks and the RASSOR rover engaging in excavation operations on the lunar surface. In both these scenarios, we use CRM terramechanics to simulate high fidelity terrain deformation, and a Chrono::Sensor camera to generate the synthetic images. Figures \ref{fig:rassor_normal} and \ref{fig:viper_normal} contain snapshots from these simulations, rendered under idealized lighting conditions so as to demonstrate the captured deformation of the terrain clearly.

\begin{figure}[!t]
    \centering
    \includegraphics[width=1.0\columnwidth]{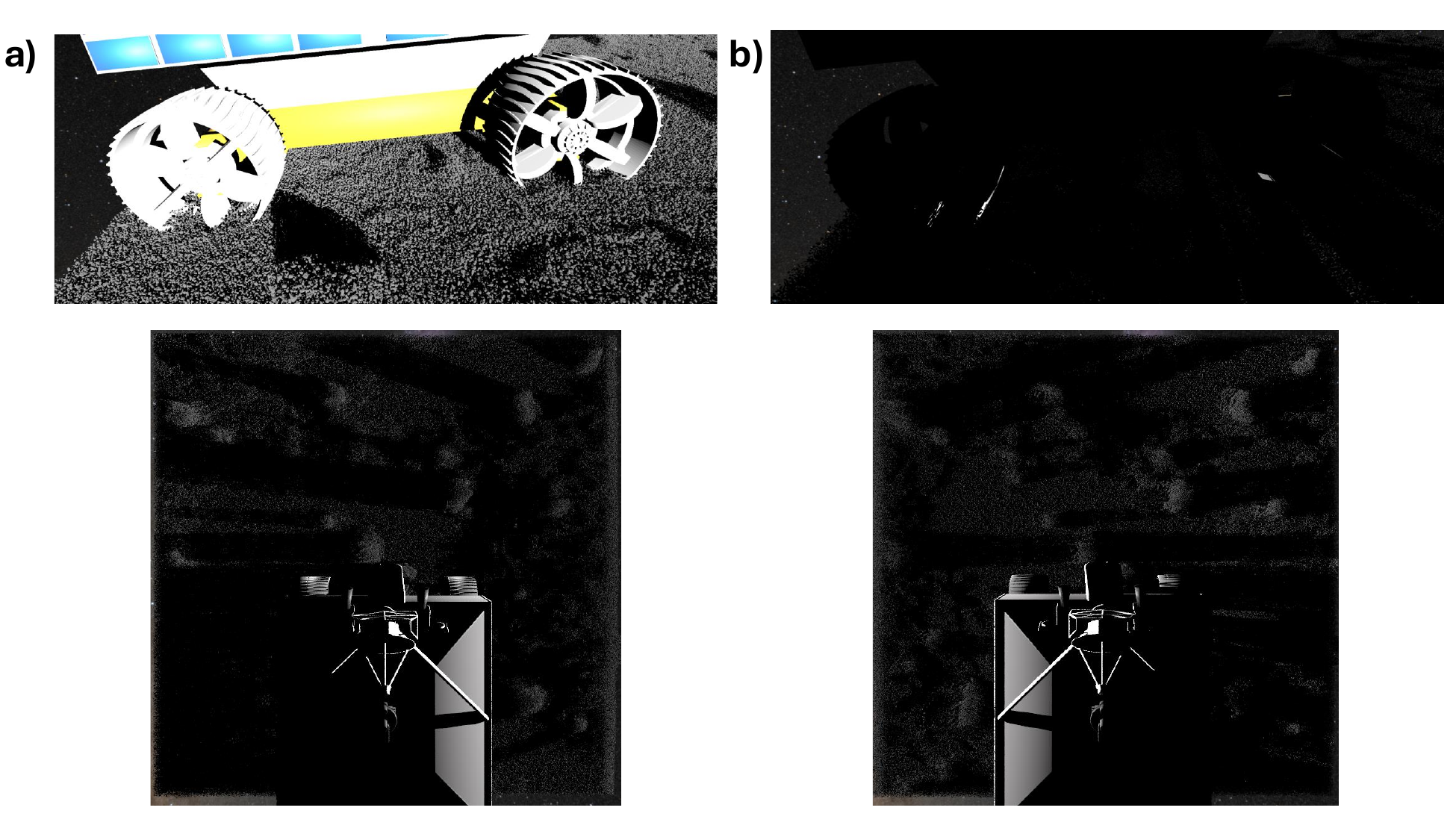}
    \caption{View from the wheel camera (top) and a Birds-eye-view (bottom) of Viper rover with realistic lunar lighting conditions.  \textbf{a)} Sun at an oblique angle directly behind the camera, \textbf{b)} Sun at an oblique angle directly opposite the camera.}
    \label{fig:viper_realistic}
\end{figure}

\begin{figure}[!t]
    \centering
    \includegraphics[width=1.0\columnwidth]{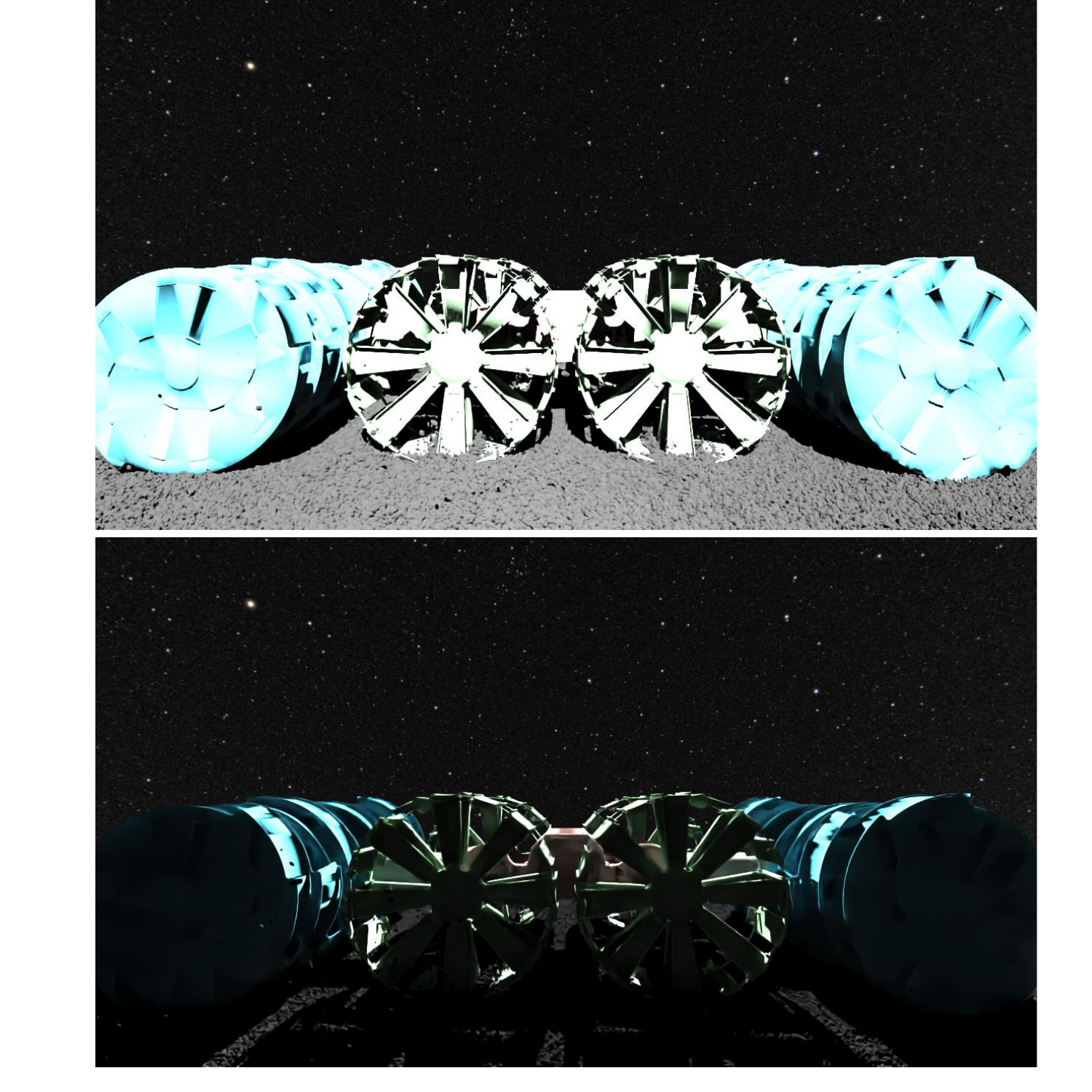}
    \caption{View from the wheel camera of the RASSOR rover with realistic lunar lighting conditions. \textbf{a)} Sun at an oblique angle directly behind the camera, \textbf{b)} Sun at an oblique angle directly opposite the camera.}
    \label{fig:rassor_realistic}
\end{figure}

Figures \ref{fig:viper_realistic} and \ref{fig:rassor_realistic} show the VIPER and RASSOR operating under realistic lunar lighting conditions. Here, we use the Hapke BRDF function \cite{Hapke2012} to model the reflection of light at the regolith surface, and also increased the intensity of the light source of the ``sun'' to simulate the harsh lighting conditions on the lunar surface. We demonstrate views from both when the sun is behind and opposite of the camera. Note that due to the fine voxelization (one voxel per particle in this case) of the terrain, we get very fine details of shadows cast by the microgeometry of the surface. Normally, in a computer graphics pipeline, these effects would be modeled using normal and shadow maps, but due to the granularity of the CRM simulation, we implicitly model these effects caused by the microgeometry of the surface.

\begin{figure}[!t]
    \centering
    \includegraphics[width=1.0\columnwidth]{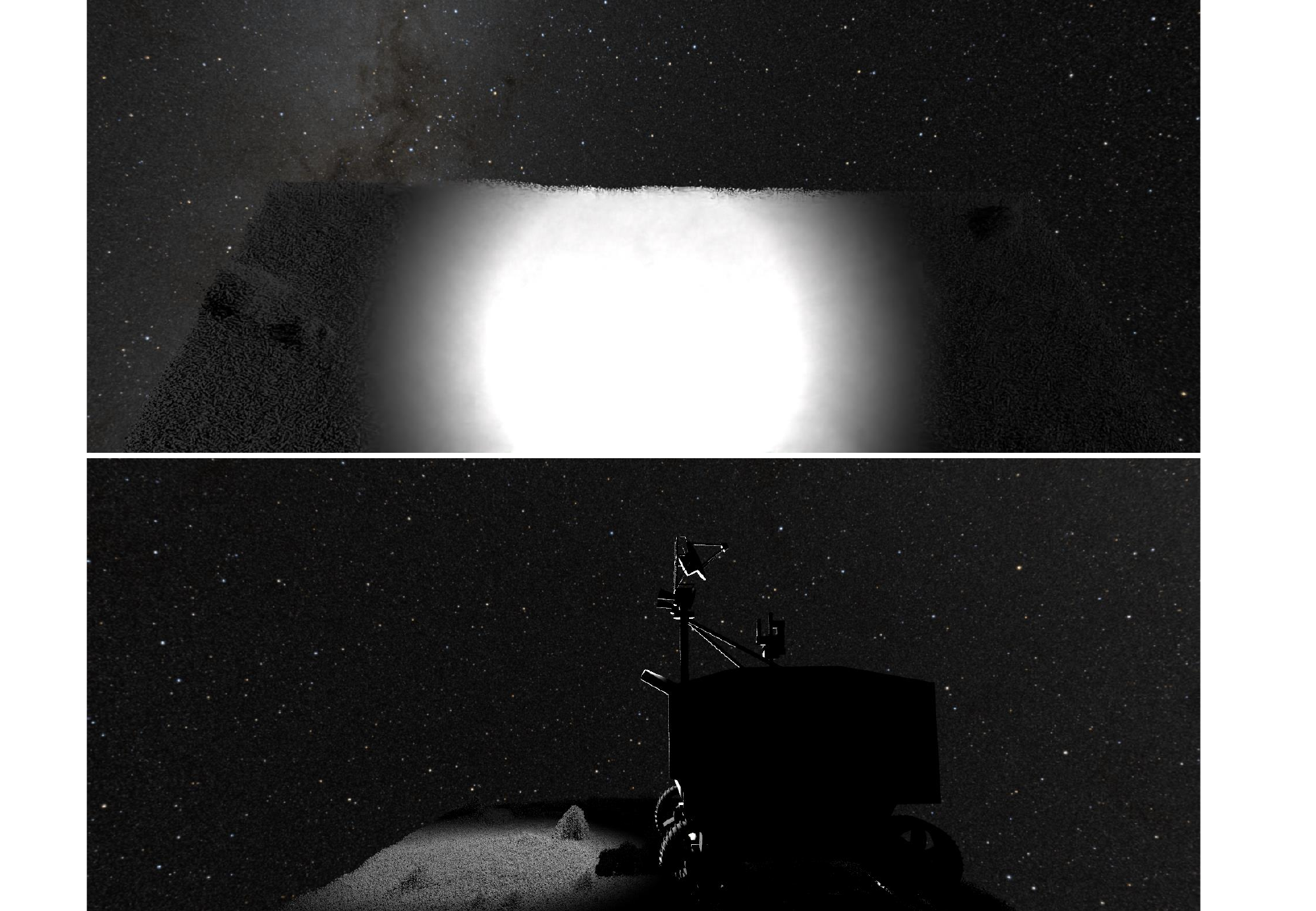}
    \caption{View from the front camera (top) and a Birds-eye-view (bottom) of the Viper rover operating under a spotlight.}
    \label{fig:viper_spot}
\end{figure}

The simulator has the capability to add spotlights to model the illumination by artificial light sources attached to rovers. Figure \ref{fig:viper_spot} shows the VIPER rover operating with a spotlight. We can see in Fig.~\ref{fig:viper_spot} a), all the terrain details are completely washed out by the spotlight. This is because of the intense opposition effect caused by the Hapke model when the sensor (front camera) and the light source (spotlight) are in the same direction.

% \sbelNoIndentTitle{Dust Model in Lunar Environments}

%% file: sections/experiments_bohsun.tex
%\sbelComment{Bo-Hsun's experiments}.

\sbelNoIndentTitle{Case Study 2: Evaluation of the Sim-to-Real Gap by Means of a Stereo Camera Algorithm}

In this experiment, a stereo camera algorithm is used to assess the sim-to-real gap between real photos and synthetic images, serving as a means to evaluate the performance of Chrono::Sensor in synthesizing lunar images within a simulation. This also demonstrates how Chrono::Sensor can aid in developing perception algorithms for use in lunar environments through simulation.

Stereo cameras can be used for 3D scene reconstruction and applied for higher level of robotic sensing tasks, such as obstacle avoidance or SLAM. The process of 3D scene reconstruction is based on the depth map that the stereo camera generates. The depth map is derived from the disparity map between the left and right cameras, where the formula is
\begin{equation}
	\label{eq:disp_to_depth}
	\textit{depth}_{ij} = \frac{b \cdot f_x}{\textit{disparity}_{ij}} \; .
\end{equation}
Above, $b$ is the baseline distance between the left and right cameras, $f_x$ is the focal length in pixels in the X direction , and \textit{disparity} is the distance in pixels of the corresponding point in the left and right photos, \textit{depth} is the distance from the scene point to the left camera, and $ij$ are the pixel indices. Simulating a good stereo camera amounts to producing an accurate disparity map.

\sbelNoIndentTitle{Datasets.} Real photos in the POLAR dataset \cite{wong2017polar} provide the target domain. It is noted that the POLAR dataset is originally used to validate stereo camera algorithms in a lunar sensing environment. Although generated on the Earth, it offers a good baseline of real data and detailed configuration for the left and right photos, as well as the ground truth depth maps. The synthetic images are generated using Chrono::Sensor, which are rendered by the Principled BRDF (the default BRDF in Chrono::Sensor, modified from the Disney model \cite{burley2012physically}), based on meshes in POLAR3D \cite{chen2023polar3d}. Notice that the virtual camera here uses an ideal pinhole camera model, since we do not have the real camera, which was used for taking the photos in the POLAR dataset, for calibrating the model parameters in the new camera model. Different exposures are corrected by assuming the CRF as a sigmoid function and conducting linear regression in the irradiance domain.

\sbelNoIndentTitle{Algorithm.} The learning-based Iterative Geometry Encoding Volume (IGEV) algorithm \cite{xu2023iterative} is selected as the stereo matching algorithm, acting as the ``judge'' in comparing the quality of real and synthetic images. This methodology provides a quantitative measure of the sim-to-real gap. IGEV uses a combined geometry encoding volume that encompasses geometry, context information, and local matching details, with leveraging a recurrent neural network to iteratively optimize the output disparity map. IGEV is trained based on pairs of left and right images from a stereo camera as inputs and outputs disparity maps under supervised training. However, the available ground truth (GT) data in both the POLAR and POLAR3D datasets are depth maps. Therefore, assuming that the real and synthetic datasets share identical stereo camera settings (i.e., baseline distance and focal length), we treat disparity maps as an encoding of depth maps and define the metric for the sim-to-real gap based on depth maps, rather than typical disparity maps.

\sbelNoIndentTitle{Metric.} The validation approach leverages the instance performance difference (IPD) concept from \cite{TR-2023-13}. For the performance metric, we use the \textit{Depth Error Ratio (DER)} in percentage (\%), defined as:
\begin{equation}
	\label{eq:depth_error_ratio}
	DER := \frac{\lvert \textit{pred\_depth\_map} - \textit{gt\_depth\_map} \rvert}{\textit{gt\_depth\_map}} \times 100\% \; ,
\end{equation}
where \textit{pred\_depth\_map} and \textit{gt\_depth\_map} are predicted and GT depth maps, respectively. DER can be used to generate error maps for visualizing the pixel-wise prediction accuracy of the trained algorithm, or be averaged over an entire image to obtain a global performance metric. Additionally, using our manually-labeled GT bounding boxes in the POLAR3D dataset for the GT real photos, DER can also be averaged over individual rocks to compute a rock-wise metric. A threshold of 5\% is widely recognized as an acceptable error margin, which we use to determine whether a rock's depth is predicted successfully in the subsequent paragraphs.

Leveraging the data in POLAR3D, the paired GT images and labeled rocks in the real and synthetic datasets allow for direct paired comparison, image-by-image or rock-by-rock, enabling us to assess the sim-to-real gap using the IGEV algorithm and the DER metric.

\sbelNoIndentTitle{Mixture of Real and Synthetic Training datasets.} Several papers have shown that training perception algorithms on mixture of real photos and synthetic images can achieve comparably good or even better performance than training on all real photos \cite{richter2016playing,ros2016synthia,peng2015learning}. The purpose of our experiment is to assess the net effect on IGEV's performance when using different mixture ratios of real and synthetic images in the algorithm's \textit{training}. We try to find the lowest required ratio of real photos in the training set to achieve comparable performance as trained on all real photos, manifesting the advantages of using the synthetic images due to the difficulty of obtaining real lunar terrain photos. We would like to see that a small amount of real images suffices to train a good IGEV algorithm. In the experiment, we created training sets with seven different reality-synthetic mixture ratios: 100\%-0\%, 83\%-17\%, 67\%-33\%, 50\%-50\%, 33\%-67\%, 17\%-83\%, and 0\%-100\%, where the first value is the percentage of real photos, and the second value is the percentage of synthetic images used in training.

\sbelNoIndentTitle{Training and Testing Details.} Of the 12 POLAR terrains, in this experiment we only used Terrains 1, 4, and 11. All pairs of left and right photos in the real dataset have their corresponding counterparts in the synthetic dataset with identical scene configurations.

First, the photo pairs were split into training and testing sets with a 2:1 ratio, and the same splitting way was applied to the synthetic dataset. Then, seven different mixture rates of the real and synthetic training sets were used to train IGEV. We used pretrained IGEV model weights downloaded from the official repository (pretrained on the Scene Flow and Middlebury datasets), and subsequently fine-tuned the model for 200 epochs for each training set. The trained model was then tested on the real and synthetic testing sets. For evaluation, we used the mean of DER values over each individual rock, denoted as $DER_{mean}$, to measure the performance value. In total, 2311 rocks were evaluated. We also assessed the proportion of rocks with $DER_{mean}$ less than 5\% as the success rate, under different training conditions.

\sbelNoIndentTitle{Evaluation Results and Discussion.} Comparison results of the averaged $DER_{mean}$ for all the rocks in the real and Principled-synthetic testing sets, predicted by IGEV models trained on different mixed training sets, are shown in Table~\ref{table:stereo_table}. When IGEV was fine-tuned without any synthetic images, the trained model performed relatively poorly on the synthetic testing set. However, after including even a small number of synthetic images in the training set, i.e., 17\%, the model's performance on the synthetic testing set improved significantly; see Table~\ref{table:stereo_table}. As the proportion of synthetic images in the training set increased, the prediction performance on the synthetic testing set improved, while performance on the real testing set gradually decreased. Nonetheless, this decrease was small. Even with only 13\% of real photos in the training set, the trained model still performed well on the real testing set, achieving an average $DER_{mean}$ of 1.520\%. However, when the training set consisted solely of synthetic images, the model's performance on the real testing set degraded significantly. It can also be observed that models trained only with real images predicted synthetic images better than models trained exclusively with synthetic images predicted real images. This suggests that predicting on synthetic images is somewhat easier than predicting on real photos.

\begin{table*}[!t]
	\caption{Averaged $DER_{mean}$ values of POLAR rocks and POLAR3D rocks. POLAR3D rocks rendered with the Principled BRDF. Values reported in percent error.}
	\label{table:stereo_table}
	\centering
	\begin{tblr}{
			colspec={lccccccc},
			row{1}={font=\bfseries},
			column{1}={font=\itshape},
			row{even}={bg=gray!10},
		}
		POLAR vs POLAR3D content $\longrightarrow$ & 100-0 & 83-17 & 67-33 & 50-50 & 33-67 & 17-83 & 0-100 \\
		\toprule
		$DER_{mean}$, POLAR rocks [\%] & 1.176 & 1.107 & 1.215 & 1.335 & 1.449 & 1.520 & 29.167 \\
		$DER_{mean}$, POLAR3D rocks [\%] & 5.952 & 1.051 & 0.866 & 0.803 & 0.705 & 0.667 & 0.677 \\
		\bottomrule
	\end{tblr}
\end{table*}
%\begin{table*}[!t]
%	\centering
%	\caption{Comparison of averaged $DER_{mean}$ of rocks in real and Principle testing set.}
%	\vspace{-12px}
%	\begin{tabular}{|c|c|c|c|c|c|c|c|}
%		\hline
%		real-Principle portion & 100-0 & 83-17 & 67-33 & 50-50 & 33-67 & 17-83 & 0-100 \\
%		\hline
%		averaged $DER_{mean}$ of real rocks [\%] & 1.176 & 1.107 & 1.215 & 1.335 & 1.449 & 1.520 & 29.167 \\
%		\hline
%		averaged $DER_{mean}$ of Principle rocks [\%] & 5.952 & 1.051 & 0.866 & 0.803 & 0.705 & 0.667 & 0.677 \\
%		\hline
%	\end{tabular}
%	\label{table:stereo_table}
%\end{table*}

\begin{figure}[!t]
	\centering
	\includegraphics[width=1.0\columnwidth]{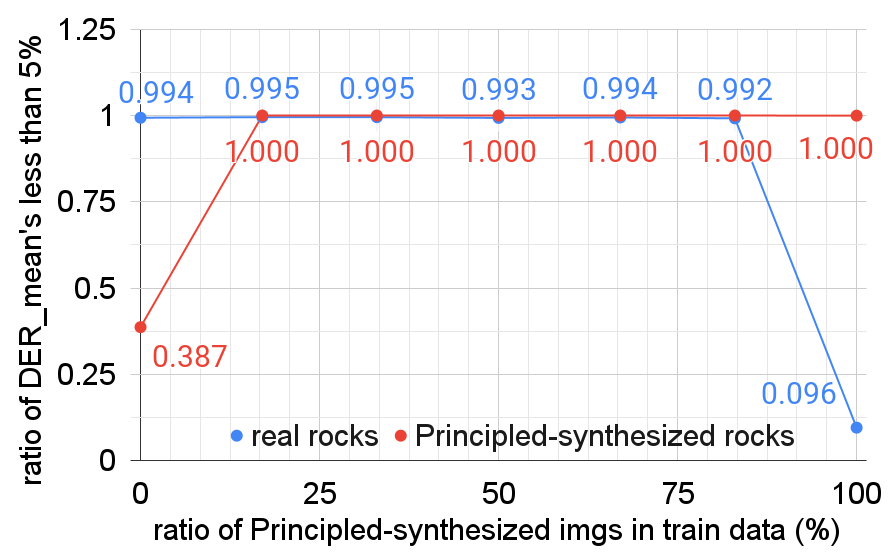}
	\caption{Comparison of the ratios of $DER_{mean}$ less than 5\% across different ratios of Principle-synthetic images in the training set.}
	\label{fig:stereo_chart}
	\vspace{-10pt}
\end{figure}

On the other hand, from the perspective of success rate, where we define success as $DER_{mean}$ being less than 5\%, Fig.~\ref{fig:stereo_chart} shows how success rates vary with different mixtures of real and synthetic pictures in the training set. The plot illustrates that even with only 17\% real images in the training set, IGEV performs well on both the real and synthetic testing sets.

\begin{figure}[!t]
	\centering
	\includegraphics[width=1.0\columnwidth]{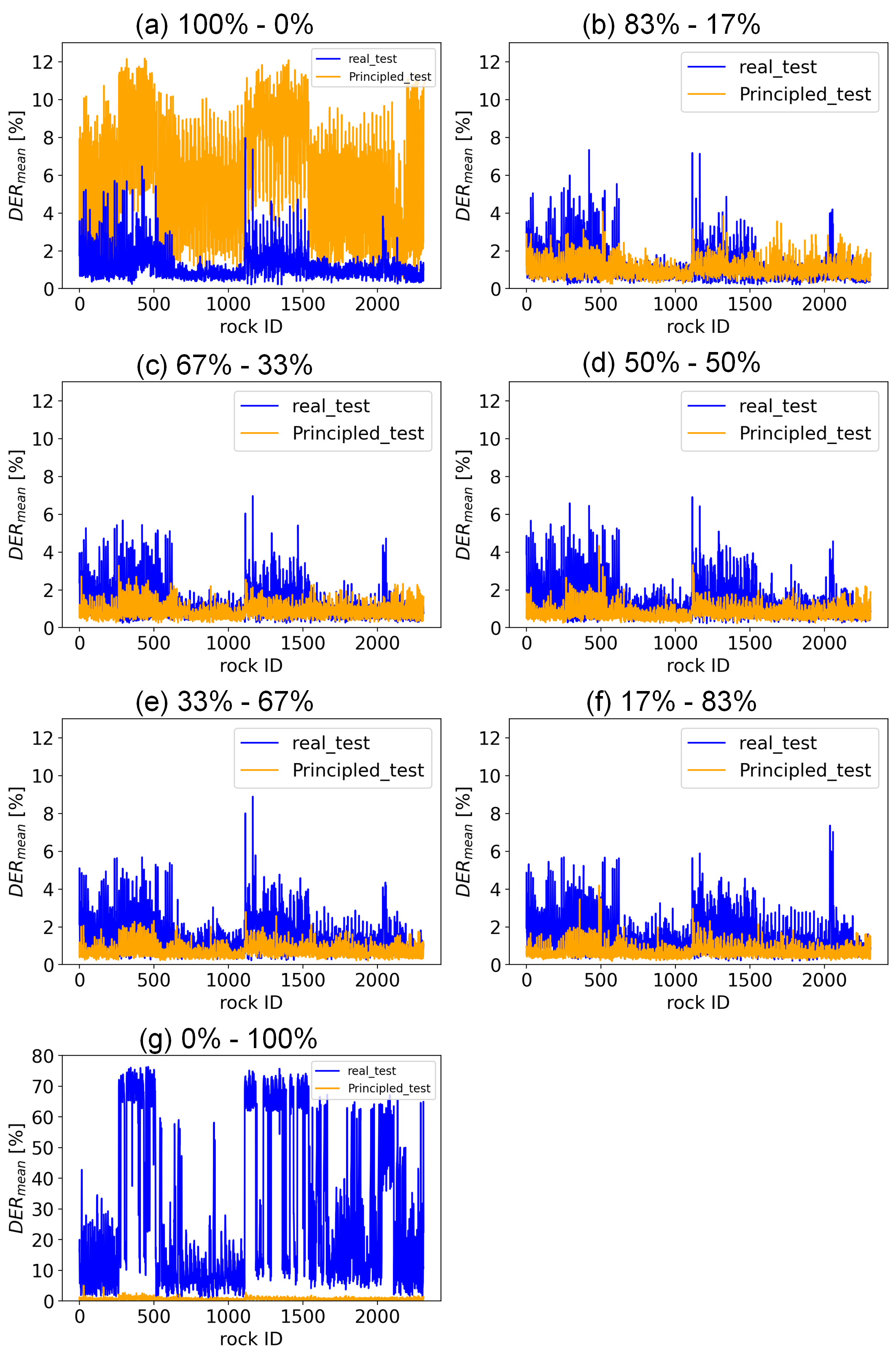}
	\caption{$DER_{mean}$ of all the rocks in the real and Principle testing sets with IGEV fin-tuned on different mixtures of real and Principle training sets.}
	\label{fig:DER_signals_plots}
\end{figure}

\begin{figure*}[!t]
	\centering
	\includegraphics[width=2.0\columnwidth]{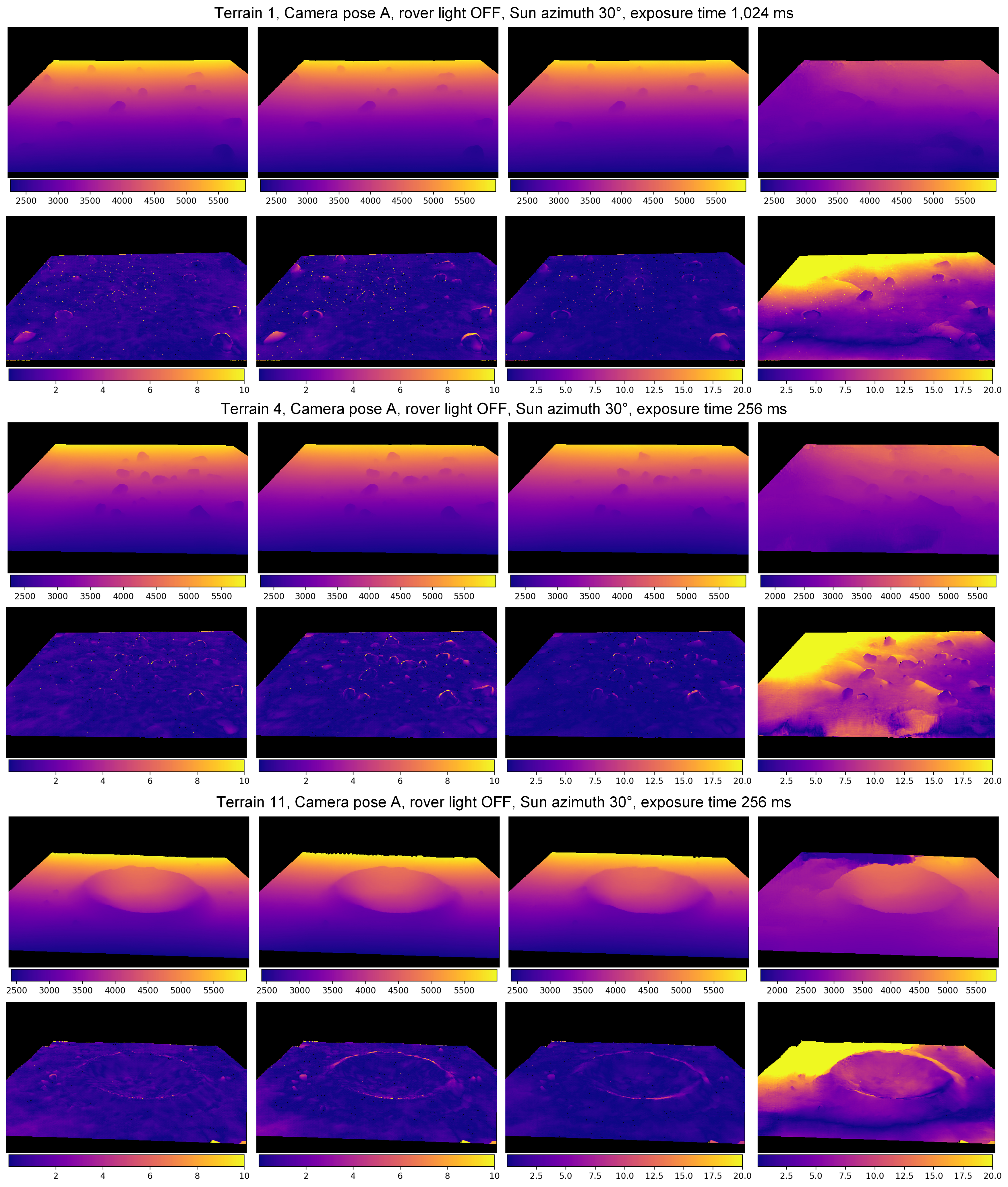}
	\caption{Comparison of depth maps and DER maps for some real photos predicted by models trained on different mixture rates of real and synthetic pictures in th training sets. Upper rows: depth maps in millimeters; lower rows: DER maps in percentages. From left to right, the reality-synthesis mixture rates are: 100\%-0\%, 50\%-50\%, 17\%-83\%, and 0\%-100\%.}
	\label{fig:stereo_demo_real}
\end{figure*}

\begin{figure*}[!t]
	\centering
	\includegraphics[width=2.0\columnwidth]{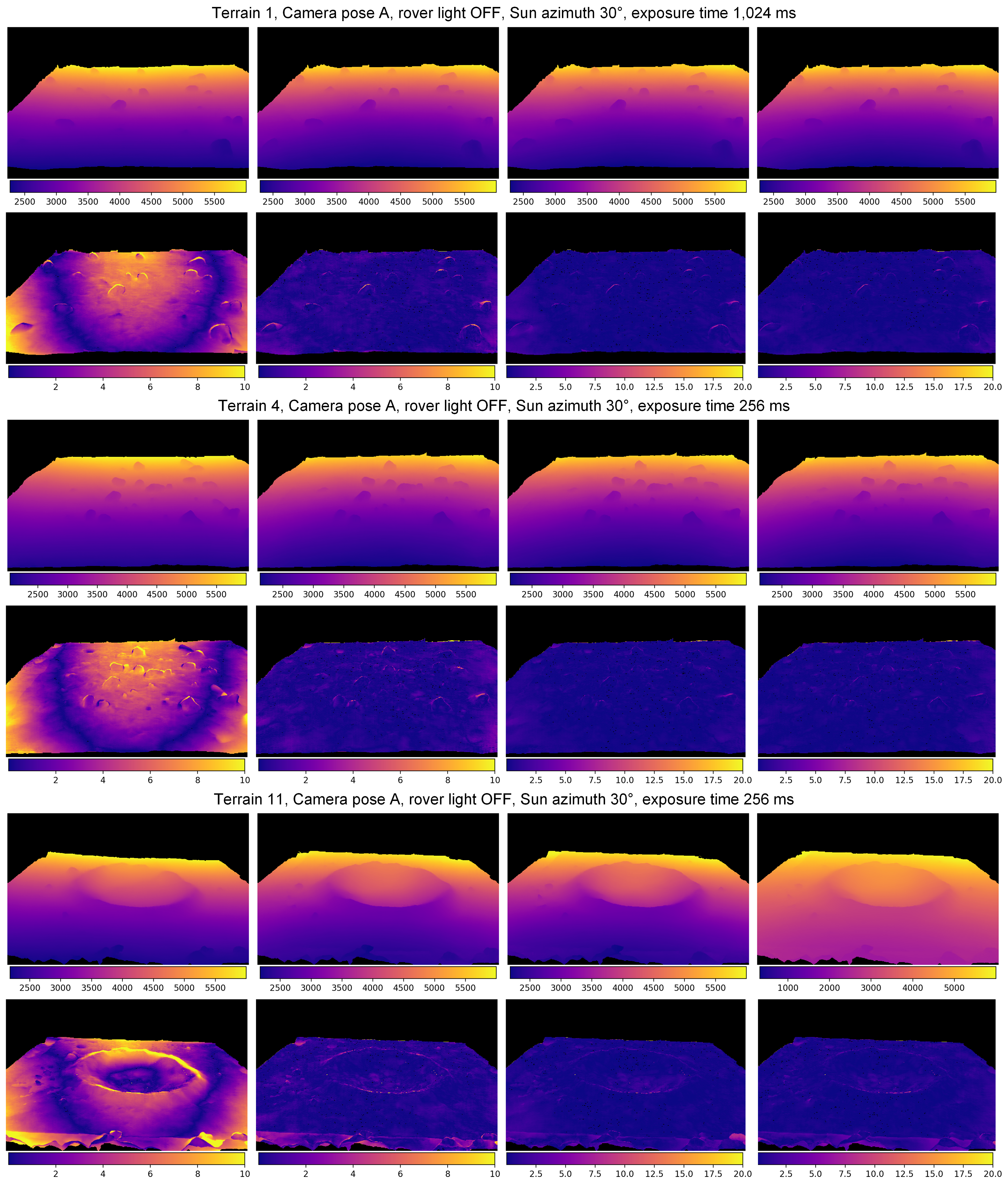}
	\caption{Comparison of depth and DER maps for some Principled-synthetic images predicted by models trained on different mixture rates of real and synthetic pictures in th training sets. Upper rows: depth maps in millimeters; lower rows: DER maps in percentages. From left to right, the reality-synthesis mixture rates are: 100\%-0\%, 50\%-50\%, 17\%-83\%, and 0\%-100\%.}
	\label{fig:stereo_demo_Disney}
\end{figure*}

Detailed comparisons of the $DER_{mean}$ values for all rocks in the real and Principled-synthetic testing sets, under different training conditions, are shown in Fig.~\ref{fig:DER_signals_plots}. The obvious differences between the real pics and synthetic images indicate that there is still a sim-to-real gap that needs to be bridged. However, the real pics and synthetic images have similar trends, showing that if the model performed well on a real rock, it also did so on the corresponding synthetic rock, and vice versa. This indicates that the synthetic images resemble well the real photos. Demonstrations of predicted depth maps and DER maps for selected real and synthetic pairs of pictures using different training set mixtures are shown in Figs.~\ref{fig:stereo_demo_real} and \ref{fig:stereo_demo_Disney}, respectively.

%% file: sections/experiments_YOLO.tex
\sbelNoIndentTitle{Case Study 3: Evaluation of the Sim-to-Real Gap by Means of an Object Detection Algorithm}

Lastly, a data-driven object detection algorithm, YOLOv5 \cite{yolov5}, is also employed to measure the sim-to-real gap between real photos and synthetic images from another perspective. In this experiment, the performance of the Principled BRDF is compared to that of the Hapke BRDF in Chrono::Sensor to evaluate which rendered images more closely resemble real photos. The comparison is conducted by training YOLOv5 on either the Principled or Hapke image datasets and testing the trained model on the real dataset, similar to the experiment in Case Study 2.

\sbelNoIndentTitle{Algorithm.} YOLOv5 is utilized for object detection. Given an input image, YOLOv5 outputs bounding boxes around the target objects in the image upon training with supervised learning. The target objects in this experiment are rocks.

\sbelNoIndentTitle{Datasets.} The same POLAR and POLAR3D datasets from the previous experiment are used. The manually labeled ground truth (GT) bounding boxes for rocks in real photos from the POLAR dataset are available in the POLAR3D dataset. For synthetic images, GT bounding boxes can be automatically generated using the ``semantic-map camera'' in Chrono::Sensor and post-processing.

\sbelNoIndentTitle{Metrics.} The evaluation metrics include the typical mean average precision (mAP) for rocks, using the intersection over union (IOU) threshold of 0.5, denoted as \textit{mAP@0.5}, and the IOU across multiple thresholds from 0.5 to 0.95, denoted as \textit{mAP@[0.5:0.95]}, respectively. In addition, for each GT bounding box of a rock, the predicted bounding box with the highest IOU is selected as the predicted label for that rock. The averaged IOU between the GT and predicted labels of all the rocks in the testing set is calculated and denoted \textit{IOU\textsubscript{mean}}. This customized metric allows for an easy rock-by-rock visualization and analysis of prediction results.

\sbelNoIndentTitle{Training and Testing Details.} The evaluation also focused only on Terrains 1, 4, and 11 of the POLAR dataset. The synthetic and real datasets were also split into training and testing sets with a 2:1 ratio. YOLOv5 was trained using transfer learning to detect rocks, starting from the pretrained YOLOv5s model weights obtained from the official website. The model was trained and validated on the training set for 200 epochs, with the best validation weights selected for testing. YOLOv5 was trained on three different training sets: real photos, Principled-synthetic images, and Hapke-synthetic images. The trained models were then cross-evaluated on the testing sets of real photos, Principled-synthetic images, and Hapke-synthetic images, respectively.

\begin{figure*}[!t]
	\centering
	\includegraphics[width=2.0\columnwidth]{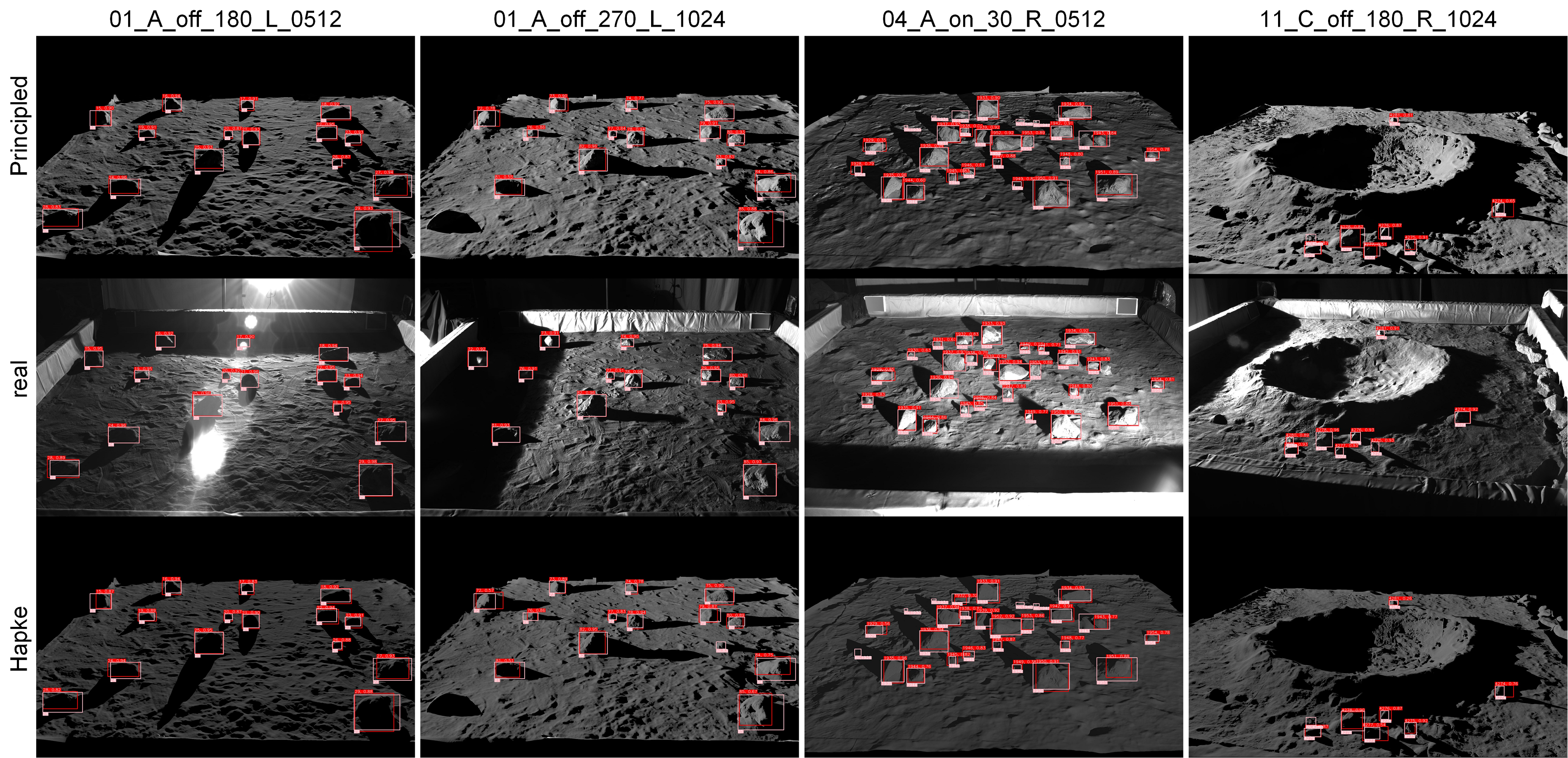}
	\caption{Illustration comparison of rock detection among real (2nd row), Principle-synthesized (1st row), and Hapke-synthesized (3rd row) images judged by YOLOv5 trained on real photos. The configuration parameters above each column are represented as: [terrain ID]\_[stereo camera position]\_[rover light status]\_[Sun azimuth]\_[(Left/ Right camera]\_[exposure time]. Pink boxes are ground-truth with rock indices, and red boxes are predictions with rock indices and prediction confidence.}
	\label{fig:yolo_result}
\end{figure*}

\sbelNoIndentTitle{Evaluation of Results and Discussion.} Figure~\ref{fig:yolo_result} shows comparisons of prediction results for several real and synthetic images. Qualitatively, both the Principled and Hapke-synthesized images closely resemble real photos, successfully capturing the light and shadow structure seen in real photos. The Principled BRDF, however, rendered rock textures more expressively, as demonstrated with Rock 85 in the second column. In contrast, the Hapke model tends to lose texture on rock surfaces when the Sun and camera are in the same direction, a phenomenon known as the \textit{opposition effect}. YOLOv5, trained on real photos, successfully detects most rocks in synthetic images, highlighting the similarity between the synthetic and real rocks.

Quantitative results are presented in Table~\ref{table:yolo_result}. Unsurprisingly, YOLOv5 performs best when tested on the same dataset it was trained on. Notably, YOLOv5 trained on real photos detects more rocks in the Principled-synthesized images, compared to the Hapke-synthesized images. Furthermore, considering the goal of using synthetic images to train perception neural networks (NNs) for real vision detection, YOLOv5 trained on Principled-synthetic images detects 6.80\% more rocks than trained on Hapke-synthetic images as measured by mAP@[0.5:0.95]. These indicate that the Principled model better fits the real photos in the POLAR dataset than the Hapke model, which is unsurprising since the images in the POLAR data set are generated on Earth, using sand instead of regolith.

\begin{table*}[!t]
	\caption{Comparison of YOLOv5 rock detection performance results.}
	\label{table:yolo_result}
	\centering
	\begin{tblr}{
			colspec={lccc},
			row{1}={font=\bfseries},
			column{1}={font=\itshape},
			row{even}={bg=gray!10},
		}
		Train\textbackslash Test & Real & Principled & Hapke \\
		\toprule
		Real      & \textbf{0.975 / 0.764 / 0.853} & 0.650 / 0.306 / 0.590 & 0.560 / 0.253 / 0.532 \\
		Principled & 0.651 / 0.330 / 0.538 & \textbf{0.913 / 0.818 / 0.899} & 0.832 / 0.670 / 0.854 \\
		Hapke     & 0.658 / 0.309 / 0.533 & 0.907 / 0.738 / 0.878 & \textbf{0.906 / 0.800 / 0.900} \\
		\bottomrule
	\end{tblr}
\end{table*}

%% file: sections/limitations.tex
Presently, in terms of sensor modeling, Chrono::Sensor has the following limitations:
\begin{itemize}[topsep=-0.5em]
	\item The most commonly used deformable terrain model in Chrono is SCM, owing to his expeditious nature. Currently, there is no support for dust simulation when one uses CRM since the phenomenological dust model employed at this time requires information that cannot be provided by SCM. 
	\item The CRM-based dust propagation model does not account for electrostatic interactions, such as the attraction of particles due to surface charge or polarization effects. 
	\item Presently, the simulation platform has no support for programmatic generation of terrain. We are currently considering open source solutions to address the limitation.
	\item At this time, there is no Level of Detail (LOD) support, that would provide high detail close-up, and lower detail at a distance. The LOD support would improve the performance of the platform, allow for efficient memory usage, and preserve visual quality. While the LOD in graphics only pertains to visual assets, for terramechanics, this LOD details includes non-graphics assets pertaining to how the terrain parameters (which are matched to grids for SCM, markers for CRM, and particles for DEM) are brought into ``focus''
	%\item The Lunar Terrain Vehicle is bound to have deformable wheels -- from elastomers, piano wire, or similar advanced materials. The nonlinear finite element support in Chrono can provide a high-fidelity solution but at a high computational cost. The simulator does not support an expeditious ``deformable wheel'' or ``deformable tire'' solution, yet work is under way to address this gap. 
	\item Currently, the ray tracing and rendering approach is built on NVIDIA's OptiX library, which requires the user to have an NVIDIA GPU. It would be desirable to implement a solution that also supports open-source alternatives such as Embree \cite{embreeSDK} and OSPRay \cite{osprayIntel2016}, which run on x86-64 platforms (Intel and AMD).
\end{itemize}

%% file: sections/conclusions.tex
We present an in-house developed, open source, publicly available simulation infrastructure that combines sensing, terramechanics, and systems dynamics to enable the in-silico analysis of lunar mobility and construction scenarios. The simulator can enable mechanical, perception, and autonomy design studies. The highlight of the contribution is the passive light sensing simulation, where the approach is anchored by ray-tracing PBR. A Hapke-based BRDF has been implemented to capture more accurately the artifacts associated with camera sensing on the Moon. Since the solution developed does not rely on third party rendering provided by Unity, Unreal Engine, or similar utilities, we have full control over the image synthesis process. As such, the camera model can be adapted to account for: noise in the image; row/column fixed pattern noise; blur due to lack or focus or fast movement; lens flare; vignetting; hot pixels; dead pixels; lens distortion; exposure; glare; compression artifacts; dust spots. The simulator can capture operations such as mobility over deformable terrains; digging; bulldozing; berming; grading. The simulator uses OpenMP parallel computing for SCM terrain, and GPU computing for CRM and DEM terramechanics. The sensing, for passive and active light, relies on ray tracing, and leverages GPU computing via NVIDIA's OptiX library, which is freely available for use under a royalty-free license. The speed of the simulation depends on the complexity of the scenario analyzed. When two or three vehicles operate on SCM or rigid terrain with long wavelength obstacle, the simulation runs in real time. When employing CRM terramechanics and sensing, the simulation runs at real time factors of tens to hundreds, and depends heavily on the GPU cards employed. Finally, DEM terramechanics elicits real time factors of hundreds to thousands, since the dynamics engine traces all the contact events at play in the granular material.

This simulation platform is complex and requires familiarity with several application areas. As such, starting from scratch requires a steep learning curve. Improving the usability and user experience represent high priorities. Other high priority is the continuous testing and validation of the infrastructure, at a time when tools like this will called upon more often than before owing to renewed interest in lunar exploration in particular, and landing on other moons and asteroids in general. Several other aspects rank high on our priority list: looking into a CPU-based sensing solution, which would relax the GPU hardware requirement; better support and subsequent validation of the lunar dust simulation approach; speed-of-execution improvements; and, most importantly, a procedural way of generating and handling at run-time large swaths of lunar terrains.